\documentclass[runningheads]{llncs}

\usepackage{amsmath,amsfonts,bm}

\def\eqref#1{equation~\ref{#1}}

\def\1{\bm{1}}

\DeclareMathAlphabet{\mathsfit}{\encodingdefault}{\sfdefault}{m}{sl}
\SetMathAlphabet{\mathsfit}{bold}{\encodingdefault}{\sfdefault}{bx}{n}

\usepackage[utf8]{inputenc} 
\usepackage[T1]{fontenc}
\usepackage{hyperref}
\usepackage{url}
\usepackage{float}
\usepackage{booktabs} 
\usepackage{subcaption} 
\usepackage{graphicx} 
\usepackage{array}
\usepackage{enumitem}
\usepackage{color}
\usepackage{multirow}

\usepackage[font={small}]{caption}

\title{Split Batch Normalization: Improving Semi-Supervised Learning under Domain Shift}

\author{Michał Zając\inst{1, 2} \and
Konrad \.Zo\l{}na\inst{1} \and
Stanisław Jastrzębski\inst{1,3}}
\authorrunning{M. Zając et al.}
\titlerunning{Split Batch Normalization}

\institute{Jagiellonian University, Krakow, Poland \and
Nomagic, Warsaw, Poland \and
New York University, New York, USA}

\begin{document}

\maketitle

\begin{abstract}
Recent work has shown that using unlabeled data in semi-supervised learning is not always beneficial and can even hurt generalization, especially when there is a class mismatch between the unlabeled and labeled examples. We investigate this phenomenon for image classification on the CIFAR-10 and the ImageNet datasets, and with many other forms of domain shifts applied (e.g.\ salt-and-pepper noise). Our main contribution is Split Batch Normalization (Split-BN), a technique to improve SSL when the additional unlabeled data comes from a shifted distribution. We achieve it by using separate batch normalization statistics for unlabeled examples. Due to its simplicity, we recommend it as a standard practice. Finally, we analyse how domain shift affects the SSL training process. In particular, we find that during training the statistics of hidden activations in late layers become markedly different between the unlabeled and the labeled examples.
\end{abstract}

\section{Introduction}

Deep neural networks (DNNs) rely on large amounts of labeled data to achieve state of the art performance on supervised learning problems such as image classification or speech recognition~\cite{Goodfellow-et-al-2016}. However, labeling data can be prohibitively costly. Consequently, a common research theme is leveraging unlabeled data to improve sample-efficiency of deep networks. 

Semi-supervised learning (SSL) is one possible approach~\cite{Chapelle2010}. SSL methods can boost performance of DNNs by jointly training on the labeled and unlabeled data~\cite{Miyato2018VirtualAT,tarvainen2017}. However, recent work has questioned the efficacy of SSL methods~\cite{oliver2018}. One of the key claims in~\cite{oliver2018} is:
 \begin{quote}
 Performance of SSL techniques can degrade drastically when the unlabeled data contains a different
 distribution of classes than the labeled data.
 \end{quote}

In this work we further investigate and improve performance of SSL techniques in the scenario when unlabeled and labeled examples do not belong to the same classes, or more generally do not come from the same distributions. We propose to compute batch normalization statistics separately for the unsupervised and supervised data. We experimentally study the effectiveness of this method on a substantially extended version of the setting considered in \cite{oliver2018}. We included other possible domain shifts in the unlabeled data (such as salt-and-pepper noise or a change in an image contrast, see also Figure~\ref{fig:distortions}), and experimented on the ImageNet dataset.

Finally, we analyse the proposed technique. Our experiments suggest that SSL training under a domain shift is difficult from the optimization point of view. We observe that hidden activation statistics significantly differ between the unlabeled and the labeled examples. We also found that in the domain shift scenario it is difficult to effectively minimize the auxiliary (unsupervised) loss term.

\section{Related Work}

Semi-supervised learning is a popular technique to leverage unlabeled data along with (typically a small amount of) labeled data~\cite{Chapelle2010}. However, in some cases using unlabeled examples can hurt performance of the model~\cite{Cozman2003,oliver2018}.

While there are many approaches to SSL, most of them implicitely or explicitely assume the unlabeled examples follow the distribution of the supervised dataset \cite{liu2018robust}. As highlighted by \cite{oliver2018}, a setting in which unlabeled examples follow a different distribution is heavily under-researched. Some works, for example \cite{laine2016temporal}, consider in their experiments unlabeled data that is out-of-distribution but do not address or analyze the problem. 

The most related works are \cite{oliver2018}, which observes that SSL performance can degrade substantially when the unlabeled dataset contains out-of-distribution examples, and \cite{liu2018robust}, which similarly to us analyzes robustness of SSL techniques. \cite{liu2018robust} considers a scenario in which there is a domain shift coming from labels that are missing not completely at random, which can lead to a mismatch in feature distribution. Importantly, the setting we consider is more general -- we allow for a systematic domain shift independent of the labeling process. Further, our Split-BN is complementary to the approach of \cite{liu2018robust}. 

Recently \cite{kalayeh2019} and \cite{deecke2019} investigated batch normalization variants that internally model the data using a multimodal distribution. In particular, \cite{deecke2019} has demonstrated improvements for multi-task supervised learning, a setting related to ours. We believe that it may be interesting direction to apply their method to SSL with domain shift.

The setting considered in our paper is also related to the field of \emph{domain adaptation} that considers different train and test data distributions \cite{ben2010theory,ganin2016domain,Li2016}. Semi-supervised methods were used recently to improve domain adaptation~\cite{Chen2018,ruder2018strong}. In particular, \cite{Chen2018} improves robustness to a domain shift using adversarial training on unlabeled examples.

Our work is also related to the recent studies of robustness of neural networks to simple statistics in the data. \cite{arpit2017closer,nasim2018} show that neural networks prioritize learning \emph{simple patterns} from the data. Another work found that convolutional neural network are highly sensitive to the high frequencies in the image space \cite{jo2017}.  Similarly, it can be shown that convolutional neural networks are overly sensitive to textures in the image \cite{wang2018learning,geirhos2018imagenettrained}.

\section{Split Batch-Normalization}

In this section we describe our main contribution: Split Batch Normalization (Split-BN), a technique to improve SSL under domain mismatch, applicable to deep neural networks. 

\subsection{Batch Normalization}

For completeness, we start with a brief review of batch normalization~\cite{Ioffe2015}. The main idea behind batch normalization is to normalize the distribution of hidden activations $\mathbf{h}$ based on the batch statistics as follows:

\begin{equation}
\hat{\mathbf{h}} = \alpha \frac{\mathbf{h} - \mu(\mathbf{h})}{\sigma(\mathbf{h})} + \beta,
\end{equation}
where $\alpha$ and $\beta$ are learnable parameters, and $\mu(\mathbf{h})$ and $\sigma(\mathbf{h})$ are the mean and the standard deviation computed on the given batch $\mathbf{h}$, called batch normalization statistics.

Batch normalization leads to large improvements in both convergence speed and generalization performance of deep neural networks~\cite{Ioffe2015}.

\subsection{Split Batch Normalization}

Typically, during the inference batch normalization statistics are computed on the whole training dataset. However, these statistics are not accurate if the deep network is applied to examples coming from a different distribution. One possible solution to this issue is to recompute the statistics on the new dataset and allow the model to learn new $\alpha$ and $\beta$ parameters~\cite{Li2016}.

Our main contribution is introducing a related technique to semi-supervised learning. We propose to \emph{compute separately batch normalization statistics for the unsupervised and supervised dataset.} By ensuring the hidden activations have the same statistics regardless of the label presence, we aim to reduce the negative effect of a domain shift between the labeled and unlabeled examples. We will refer to this technique as Split Batch Normalization (Split-BN).

More precisely, let $\mathbf{h}_l$ and $\mathbf{h}_u$ denote the labeled and the unlabeled examples in a given batch $\mathbf{h}$, respectively. Then Split-BN normalizes the hidden activations as follows:

\begin{align}
\hat{\mathbf{h_u}} &= \alpha \frac{\mathbf{h_u} - \mu(\mathbf{h_u})}{\sigma(\mathbf{h_u})} + \beta \\
\hat{\mathbf{h_l}} &= \alpha \frac{\mathbf{h_l} - \mu(\mathbf{h_l})}{\sigma(\mathbf{h_l})} + \beta.
\end{align}

Analogously, during the inference means and standard deviations are computed separately on the labeled, and the unlabeled examples. Even though the statistics are computed independently, the $\alpha$ and $\beta$ parameters are shared\footnote{We experimentally found that not sharing $\alpha$ and $\beta$ leads to finding degenerated solutions where $\alpha$ vanishes.}.

\section{Experiments}

In this section we evaluate the effectiveness of Split-BN in semi-supervised learning under a domain shift.  

\begin{figure}
\captionsetup[subfigure]{position=top}
    \centering
    \begin{subfigure}[b]{0.23\textwidth}
        \centering
        \tiny
        \includegraphics[width=0.9\textwidth]{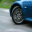}
        \caption{none}
    \end{subfigure}
    \begin{subfigure}[b]{0.23\textwidth}
        \centering
        \includegraphics[width=0.9\textwidth]{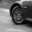}
        \caption{grayscale}
    \end{subfigure}
    \begin{subfigure}[b]{0.23\textwidth}
        \centering
        \includegraphics[width=0.9\textwidth]{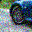}
        \caption{uniform noise}
    \end{subfigure}
    \begin{subfigure}[b]{0.23\textwidth}
        \centering
        \includegraphics[width=0.9\textwidth]{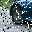}
        \caption{salt-and-pepper}
    \end{subfigure}

    \begin{subfigure}[b]{0.23\textwidth}
        \centering
        \includegraphics[width=0.9\textwidth]{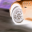}
        \caption{inverted colors}
    \end{subfigure}
    \begin{subfigure}[b]{0.23\textwidth}
        \centering
        \includegraphics[width=0.9\textwidth]{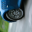}
        \caption{rotation}
    \end{subfigure}
    \begin{subfigure}[b]{0.23\textwidth}
        \centering
        \includegraphics[width=0.9\textwidth]{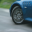}
        \caption{random contrast}
    \end{subfigure}
    \begin{subfigure}[b]{0.23\textwidth}
        \centering
        \includegraphics[width=0.9\textwidth]{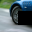}
        \caption{occlusion}
    \end{subfigure}
\caption{All distortions used in the paper to introduce a domain shift between the labeled and unlabeled data.} \label{fig:distortions}
\end{figure}

%
We first describe experimental setting in Section~\ref{sec:expset} and Section~\ref{subsec:datasets}. In Section~\ref{sec:cm} we analyse the class-mismatch setting of~\cite{oliver2018}. Next, in Section~\ref{sec:other_distortions} we study Split-BN on other forms of domain shift (see  Figure~\ref{fig:distortions}). The experiments are performed on the CIFAR-10 and the ImageNet datasets using state-of-the-art semi-supervised methods.

Our key result is that \emph{Split-BN typically cancels the performance gap between supervised learning and SSL with misaligned data; and in some cases even improves performance by a similar amount as SSL without a domain shift.}

In the final three Sections~\ref{subsec:activations_analysis}-\ref{sec:isbnculprit} we analyse how Split-BN improves performance. The experiments suggest that the negative effect of misaligned data is related to a difficulty in optimization and a difference in hidden activation statistics between the labeled and the unlabeled examples.

\subsection{Experimental setting}
\label{sec:expset}

Our experiments largely base on~\cite{oliver2018}. In particular, we use the same strong Wide ResNet model (WRN-28-2 variant with depth 28 and width 2)~\cite{Zagoruyko2016WRN}. 
In all experiments we use the following: batch size equal to 100; Adam optimizer with all hyperparameters but learning rate set to default. Every number we report is accuracy on test set computed using the checkpoint with the highest validation accuracy during the training. Every experiment is repeated 2 times. We perform experiments with two state-of-the-art SSL methods: Mean Teacher (MT) by~\cite{tarvainen2017} and Virtual Adversarial Training (VAT) by~\cite{Miyato2018VirtualAT}. In all experiments we use 400 labels per class.

\begin{table}
\caption{Test accuracy under the class-mismatch from \cite{oliver2018}, on the CIFAR-10 dataset. Columns correspond to various level of class mismatch between labeled and unlabeled data -- from $0\%$ (no mismatch) to $100\%$ (no shared classes). Our Split-BN improves upon the results reported in \cite{oliver2018}, and in particular removes most of the gap between SSL with class mismatch and pure supervised training.}
  \label{tab:cifar10_class_mismatch}
\centering
\begin{tabular}{>{\scriptsize}l|>{\scriptsize}c||>{\scriptsize}c|>{\scriptsize}c||>{\scriptsize}c|>{\scriptsize}c}
 \toprule
  & \textbf{Supervised} & \textbf{MT}  & \textbf{MT + Split-BN}  & \textbf{VAT}  & \textbf{VAT + Split-BN} \\
  \midrule
 $0\%$  & \multirow{5}{*}{77.0 $\pm$ 0.4} & \textbf{77.7 $\pm$ 0.5}  &  \textbf{77.1 $\pm$ 0.2}  &  \textbf{79.3 $\pm$ 0.3}  &  76.6 $\pm$ 0.3 \\
 $25\%$  & & 75.5 $\pm$ 0.7  &  \textbf{77.6 $\pm$ 0.1}  &  \textbf{76.3 $\pm$ 0.5}  &  \textbf{76.6 $\pm$ 0.4} \\
 $50\%$  & & 74.8 $\pm$ 0.3  &  \textbf{76.8 $\pm$ 0.1}  &  \textbf{75.7 $\pm$ 0.5}  &  \textbf{76.3 $\pm$ 1.0} \\
 $75\%$  & & 74.5 $\pm$ 0.1  &  \textbf{77.1 $\pm$ 0.4}  &  73.8 $\pm$ 0.1  &  \textbf{76.1 $\pm$ 0.0} \\
 $100\%$ & & 73.9 $\pm$ 0.4 &  \textbf{76.2 $\pm$ 0.5} &  72.6 $\pm$ 0.2 &  \textbf{76.6 $\pm$ 0.5}\\
 \bottomrule
 
\end{tabular}
\end{table}

For simplicity, we fix most of the hyperparameter values to the ones used in~\cite{oliver2018} CIFAR-10 experiments. For a more detailed description, please refer to Appendix~\ref{app:experimental_details}. The code to reproduce  the results will be released upon publication. 

\subsection{Datasets and domain shifts}
\label{subsec:datasets}

We run our experiments on the CIFAR-10, and on the ImageNet dataset. Additionally, in order to work with complex image data but with small amount of classes and number of images per class similar to that in CIFAR10, we created 8A8O-Imagenet (8 animal and 8 other classes). For details of the dataset refer to the Appendix~\ref{app:8a8o}.

In addition to the class mismatch from \cite{oliver2018}, we use popular image distortions to introduce a systematic difference between the labeled and unlabeled examples. In total, we study $8$ different domain shift scenarios: the first based on a class mismatch, and the rest based on applying a fixed type of image distortion to all images in the unlabeled dataset. These distortions are meant to represent challenge present in real world applications. 




Figure~\ref{fig:distortions} shows all seven distortions applied to a randomly chosen image. The details of the applied distortions are as follows:
\begin{enumerate}
\item grayscale: use \texttt{tf.image.rgb\_to\_grayscale} and then stack 3 times to have the same input size;
\item uniform noise: add uniform noise in range $[-0.2, 0.2]$ and then clip the values back to $[0, 1]$;
\item salt-and-pepper: every pixel becomes white with probability $10\%$, black with probability $10\%$, and stays the same otherwise;
\item inverted colors: every color value goes from $x$ to $1 - x$;
\item rotation: the whole image is rotated by $90^{\circ}$ counterclockwise;
\item random contrast: contrast is changed to a random value taken uniformly from $[0.2, 0.8]$;
\item occlusion: a black 14x14 square is placed on top of the picture at a random location.
\end{enumerate}

\begin{table}
\centering
    \caption{Test accuracy on the class-mismatch setting on the ImageNet. Columns correspond to various level of class mismatch between labeled and unlabeled data -- from $0\%$ (no mismatch) to $100\%$ (no shared classes).} 
  \label{tab:imagenet_class_mismatch}
\begin{tabular}{>{\scriptsize}l|>{\scriptsize}c||>{\scriptsize}c|>{\scriptsize}c}
 \toprule
 & \textbf{Supervised} & \textbf{MT}  & \textbf{MT + Split-BN} \\
 \midrule
 $0\%$  & \multirow{5}{*}{61.5 $\pm$ 1.2} &  \textbf{65.9 $\pm$ 0.8}  &  \textbf{65.5 $\pm$ 1.2} \\
 $25\%$  & & \textbf{67.2 $\pm$ 1.0}  &  \textbf{65.8 $\pm$ 0.9} \\
 $50\%$  & & 65.0 $\pm$ 0.6  &  \textbf{66.0 $\pm$ 0.0} \\
 $75\%$  & & \textbf{63.8 $\pm$ 1.6}  &  \textbf{65.5 $\pm$ 1.1} \\
 $100\%$ & & \textbf{64.0 $\pm$ 1.3} &  \textbf{64.1 $\pm$ 0.8} \\
 \bottomrule
\end{tabular}
\end{table}

\subsection{Class mismatch results}
\label{sec:cm}

First, we test our method on the class mismatch setting on the CIFAR-10 dataset, as discussed in \cite{oliver2018}. Table~\ref{tab:cifar10_class_mismatch} reports the results. We can see that in some cases baseline performance is even $3\%$ below accuracy achieved by pure supervised training that does not use additional unsupervised data. Critically, using Split-BN cancels most of the performance degradation in these cases. Split-BN makes the use of SSL methods safe and robust to class mismatch.

To further study the class mismatch setting, we run experiments on the more challenging ImageNet dataset~\cite{imagenet_cvpr09}. The experiment is run in two variants: we either select as training data 20 randomly classes as the supervised dataset, or we use 8A8O-ImageNet that has only animal classes in the training set (see Subsection~\ref{subsec:datasets} for details). In both cases we vary the number of unlabeled examples that do not come from the training set classes.

Results for the first variant are reported in Table~\ref{tab:imagenet_class_mismatch}. In contrast to the previous experiment, there is an apparent improvement over the pure supervised case and our Split-BN performs comparably to regular SSL. 

On the other hand, selecting only animal classes in 8A8O-Imagenet leads to a noticeable (up to $2\%$) degradation in SSL's methods performance, as shown in Table~\ref{tab:our_imagenet_class_mismatch}. This stresses that the robustness to class mismatch heavily depends on the dataset, and the relationship between the labeled and the unlabeled examples.

\begin{table}
\centering
    \caption{Test accuracy on the class-mismatch setting on the 8A8O-ImageNet. Columns correspond to various level of class mismatch between labeled and unlabeled data -- from $0\%$ (no mismatch) to $100\%$ (no shared classes). }
  \label{tab:our_imagenet_class_mismatch}
\begin{tabular}{>{\scriptsize}l|>{\scriptsize}c||>{\scriptsize}c|>{\scriptsize}c||>{\scriptsize}c|>{\scriptsize}c}
  \toprule
  & \textbf{Supervised} & \textbf{MT}  & \textbf{MT + Split-BN}  & \textbf{VAT}  & \textbf{VAT + Split-BN} \\
  \midrule
 0\% & \multirow{5}{*}{52.7 $\pm$ 0.4} & \textbf{55.7 $\pm$ 0.2} & \textbf{55.6 $\pm$ 0.5} & \textbf{52.1 $\pm$ 1.9} & \textbf{52.7 $\pm$ 0.1} \\
 25\% &  & \textbf{54.0 $\pm$ 0.3} & 52.8 $\pm$ 0.3 & 50.3 $\pm$ 0.8 & \textbf{52.7 $\pm$ 0.0} \\
 50\%  & & \textbf{52.4 $\pm$ 0.4} & \textbf{52.6 $\pm$ 0.1} & \textbf{51.1 $\pm$ 2.1} & \textbf{51.2 $\pm$ 0.2} \\
 75\% &  & 49.7 $\pm$ 0.2 & \textbf{52.1 $\pm$ 0.8} & \textbf{51.6 $\pm$ 0.4} & 50.7 $\pm$ 0.5 \\
 100\%  & & 49.1 $\pm$ 0.5 & \textbf{53.0 $\pm$ 0.4} & 49.1 $\pm$ 0.7 & \textbf{50.3 $\pm$ 0.2} \\
 \bottomrule
\end{tabular}
\end{table}

To summarize, we can observe that in the majority of cases Split-BN either improves or matches performance of vanilla batch normalization.

\subsection{Image distortions results}
\label{sec:other_distortions}
\begin{table}[bp]
    \caption{Test accuracy for each SSL technique under other than class-mismatch forms of domain shift, on the CIFAR-10 dataset. Semi-supervised learning methods are generally not robust to domain shift; in the worst case performance degrades $4\%$ in comparison to pure supervised training.  Our Split-BN in most cases leads to an improvement over both SSL and pure supervised training.}
  \label{tab:distortions}
\centering
\setlength\tabcolsep{1.5pt}
\begin{tabular}{>{\scriptsize}l|>{\scriptsize}c||>{\scriptsize}c|>{\scriptsize}c||>{\scriptsize}c|>{\scriptsize}c}
 \toprule
  & \textbf{Supervised} & \textbf{MT}  & \textbf{MT + Split-BN}  & \textbf{VAT}  & \textbf{VAT + Split-BN} \\
  \midrule
 None & \multirow{8}{*}{79.4 $\pm$ 0.1} &  \textbf{83.2 $\pm$ 0.1}  &  \textbf{83.1 $\pm$ 0.6}  &  \textbf{85.1 $\pm$ 0.0}  &  \textbf{85.2 $\pm$ 0.1} \\
 Grayscale  & &  79.3 $\pm$ 0.3  &  \textbf{80.4 $\pm$ 0.0}  &  81.4 $\pm$ 0.4  &  \textbf{83.3 $\pm$ 0.1} \\
 Uniform noise  & &  \textbf{79.4 $\pm$ 0.4}  &  \textbf{79.4 $\pm$ 0.1}  &  \textbf{79.3 $\pm$ 0.4}  &  78.2 $\pm$ 0.5 \\
 Salt-and-pepper  & &  \textbf{78.2 $\pm$ 0.4}  &  \textbf{78.8 $\pm$ 0.5}  &  \textbf{78.7 $\pm$ 0.7}  &  \textbf{79.4 $\pm$ 0.8} \\
 Inverted colors  & &  75.2 $\pm$ 0.2  &  \textbf{79.8 $\pm$ 0.6}  &  \textbf{79.0 $\pm$ 0.5}  &  \textbf{78.8 $\pm$ 0.2} \\
 Rotation $90^{\circ}$  & &  75.9 $\pm$ 0.1  &  \textbf{80.1 $\pm$ 0.1}  &  77.6 $\pm$ 0.4  &  \textbf{78.6 $\pm$ 0.0} \\
 Random contrast  & &  80.9 $\pm$ 0.0  &  \textbf{82.6 $\pm$ 0.6}  &  82.0 $\pm$ 0.6  &  \textbf{83.9 $\pm$ 0.2} \\
 Occlusion & &  78.6 $\pm$ 0.5 &  \textbf{79.9 $\pm$ 0.0} &  \textbf{78.4 $\pm$ 0.2} &  \textbf{80.0 $\pm$ 1.4} \\
 \bottomrule
\end{tabular}
\end{table}

In this section we empirically show that semi-supervised learning performance is not robust when unlabeled data is distorted as described in Section~\ref{subsec:datasets}. Then we show that Split-BN generalizes to this new setting. Table~\ref{tab:distortions} summarizes the results. In all cases with domain shift we observe a large gap in performance, typically between $1\%$ and $4\%$ accuracy drop. Hence, the phenomenon reported by \cite{oliver2018} for class mismatch is confirmed for other forms of domain shifts. Additionally, Split-BN typically cancels most of the negative effect resulting from the domain mismatch, and in many cases leads to a large improvement in performance.


Next, as in the previous section, we repeat the experiment on the 8A8O-ImageNet dataset. Results are shown in Table ~\ref{tab:distortions_our_imagenet}. Here semi-supervised learning always under-performs compared to pure supervised training. Split-BN again cancels the negative effect of domain shift, but does not provide further improvements.

\begin{table}
    \caption{Test accuracy for each SSL technique under other than class-mismatch forms of domain shift, on animal classes of 8A8O-ImageNet. Semi-supervised learning methods are generally not robust to domain shift; in the worst case performance degrades $4\%$ in comparison to pure supervised training.  Our Split-BN in most cases leads to an improvement over both SSL and pure supervised training.}
 
  \label{tab:distortions_our_imagenet}
\centering
\setlength\tabcolsep{1.5pt}
\begin{tabular}{>{\scriptsize}l|>{\scriptsize}c||>{\scriptsize}c|>{\scriptsize}c||>{\scriptsize}c|>{\scriptsize}c}
 \toprule
  & \textbf{Supervised} & \textbf{MT}  & \textbf{MT + Split-BN}  & \textbf{VAT}  & \textbf{VAT + Split-BN} \\
  \midrule
 None & \multirow{8}{*}{52.7 $\pm$ 0.4} & \textbf{55.7 $\pm$ 0.2} & \textbf{55.6 $\pm$ 0.5} & \textbf{52.1 $\pm$ 1.9} & \textbf{52.7 $\pm$ 0.1} \\
 Grayscale  & & 49.8 $\pm$ 0.3 & \textbf{52.3 $\pm$ 0.3} & 50.5 $\pm$ 1.2 & \textbf{52.9 $\pm$ 0.6} \\
 Uniform noise  & & 51.6 $\pm$
0.1 & \textbf{53.3 $\pm$ 1.5} & 50.8 $\pm$ 0.5 & \textbf{52.3 $\pm$ 0.2} \\
 Salt-and-pepper  & & \textbf{51.1 $\pm
$ 0.9} & \textbf{52.2 $\pm$ 0.8} & \textbf{52.6 $\pm$ 0.4} & \textbf{51.9 $\pm$ 0.6} \\
 Inverted colors  & & 48.9 $\pm$
0.6 & \textbf{52.0 $\pm$ 0.1} & \textbf{50.7 $\pm$ 0.6} & \textbf{51.3 $\pm$ 0.5} \\
 Rotation $90^{\circ}$ & & \textbf{51.4 $\pm$ 0.7} & \textbf{51.9 $\pm$ 0.4} & 51.3 $\pm$ 0.3 & \textbf{52.1 $\pm$ 0.3} \\
 Random contrast  & &  \textbf{50.8 $\pm
$ 0.5} & \textbf{51.7 $\pm$
0.7} & \textbf{51.7 $\pm$ 0.7} & \textbf{51.2 $\pm$ 0.7} \\
 Occlusion & &  50.3 $\pm$ 0.3 & \textbf{52.3 $\pm$
 1.0} & 50.2 $\pm$ 0.5 & \textbf{53.0 $\pm$ 0.1} \\
 \bottomrule
\end{tabular}
\end{table}

\subsection{Split-BN brings closer unlabeled and labeled examples' activations}
\label{subsec:activations_analysis}

In this section we try to shed light on how Split-BN improves performance by visualizing the distributions of hidden activations during training. We analyse one of the experiments described in Section~\ref{sec:other_distortions} corresponding to training with Mean Teacher and a domain shift introduced by a rotation. 

\begin{figure}[H]
\captionsetup[subfigure]{position=top}
    \centering
    \begin{subfigure}[b]{0.49\textwidth}
        \centering
        \tiny
        \includegraphics[width=\textwidth]{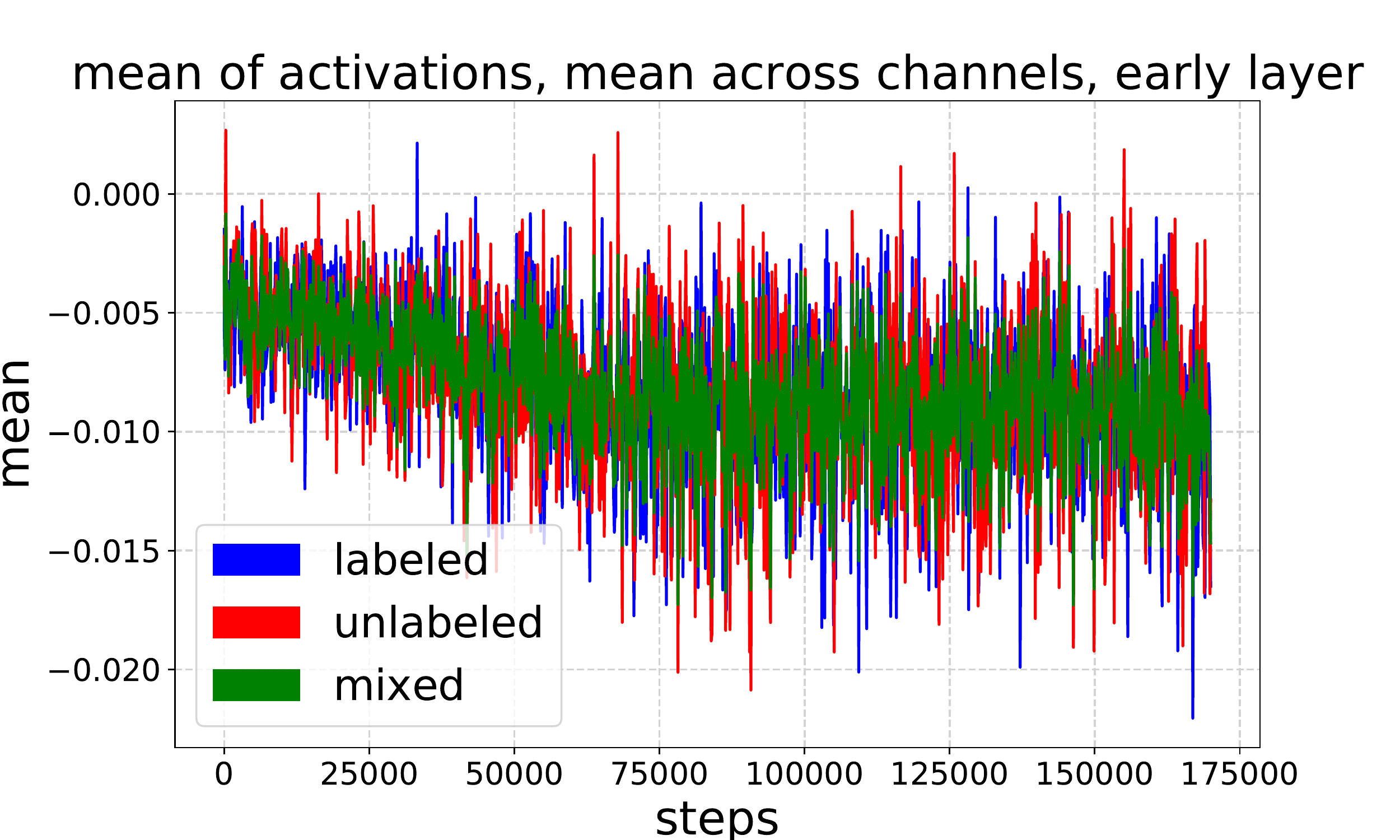}
    \end{subfigure}
    \begin{subfigure}[b]{0.49\textwidth}

        \centering
        \includegraphics[width=\textwidth]{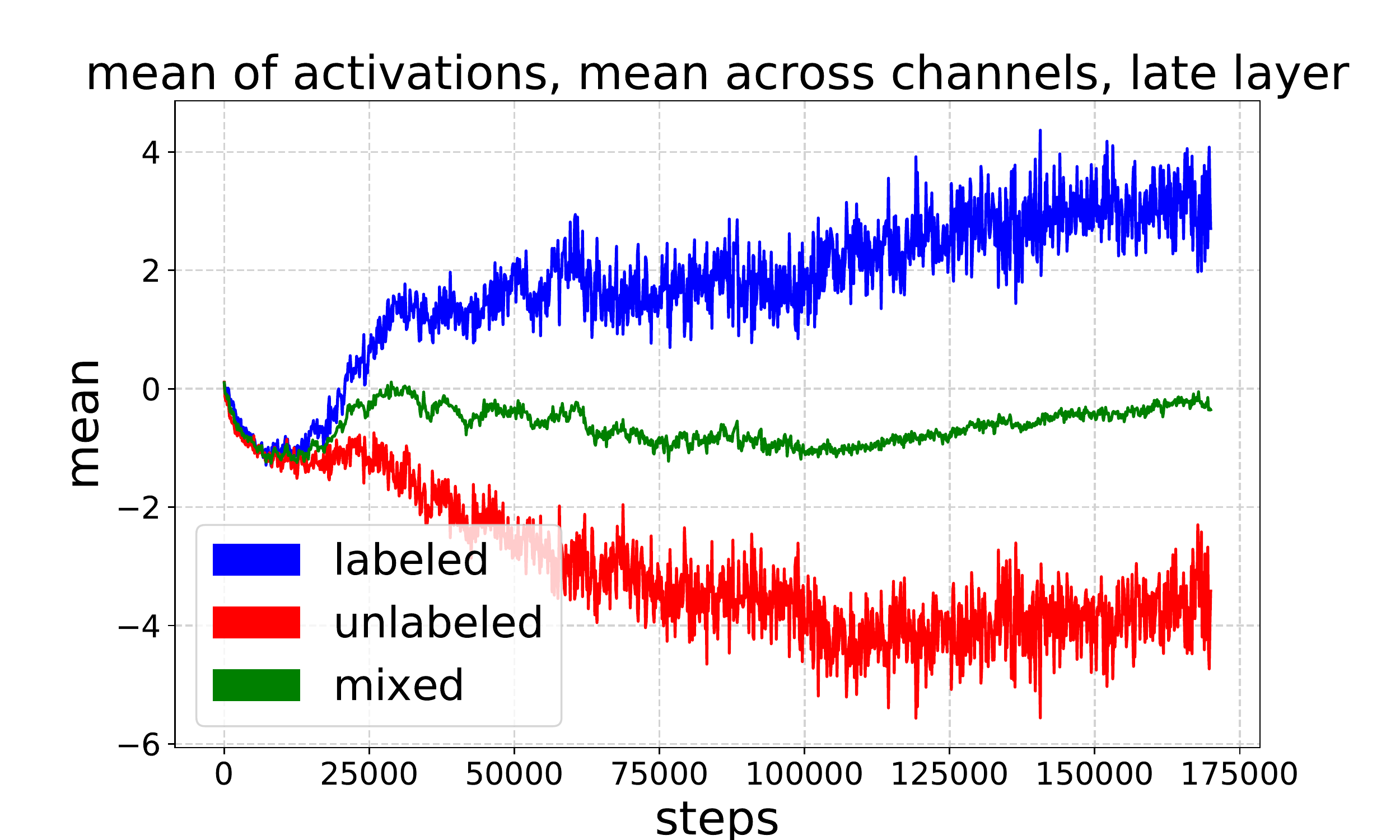}
    \end{subfigure}

     \begin{subfigure}[b]{0.49\textwidth}
         \centering
         \includegraphics[width=\textwidth]{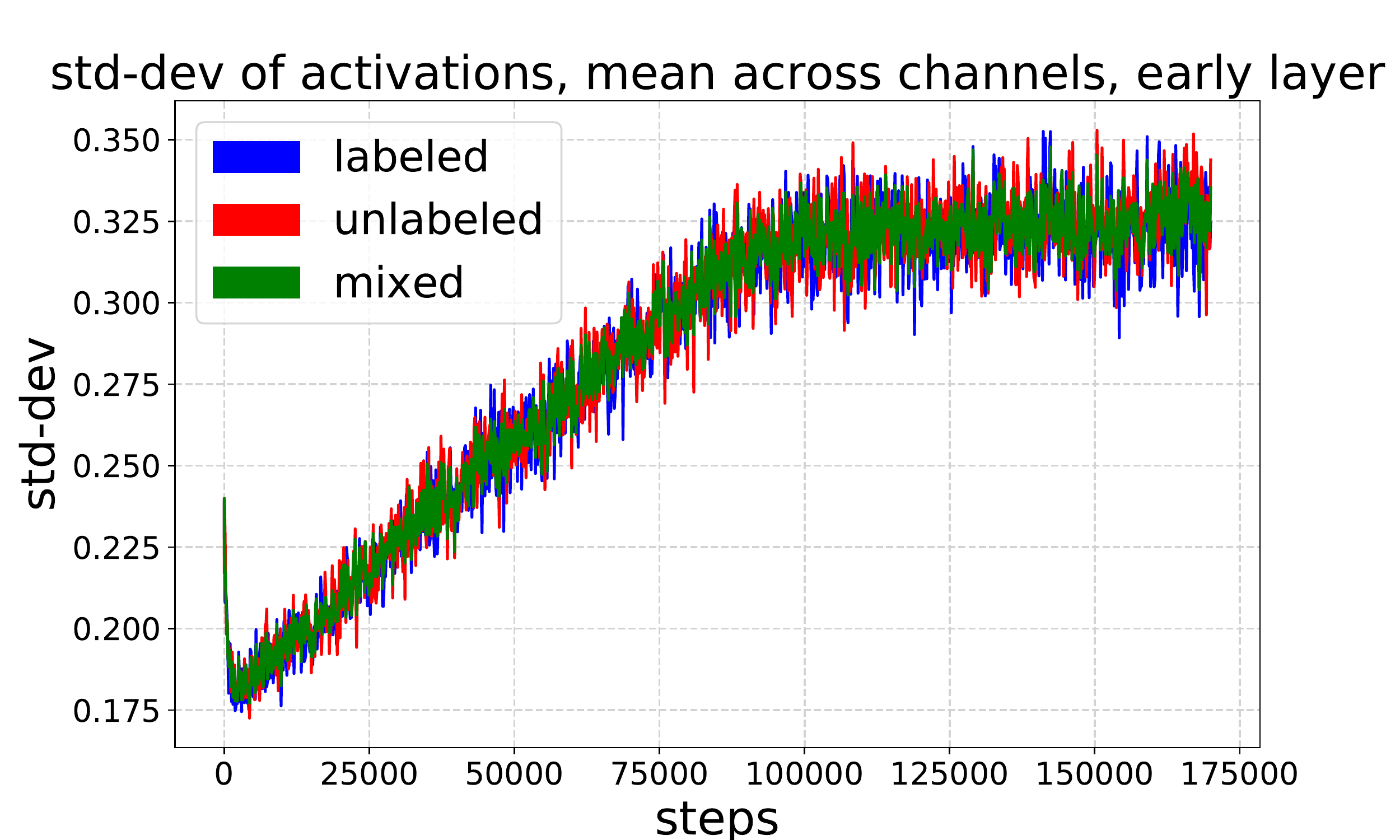}
     \end{subfigure}
     \begin{subfigure}[b]{0.49\textwidth}
         \centering
         \includegraphics[width=\textwidth]{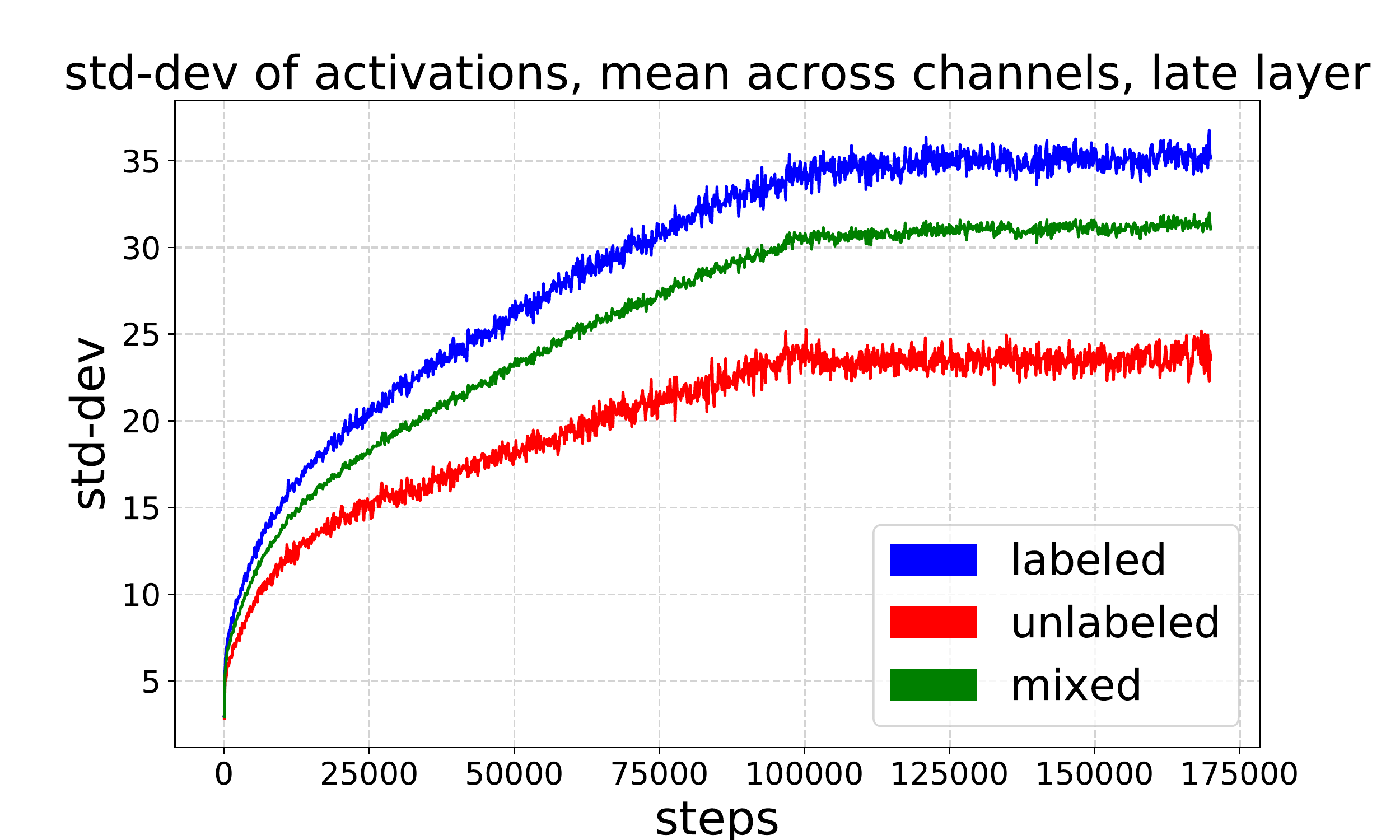}
     \end{subfigure}
    
    \caption{Means (top) and standard deviations (bottom) of activations before the first batch normalization layer (left), and the last batch normalization layer (right). Experiment run on the Wide ResNet model from \cite{oliver2018}, and the CIFAR-10 dataset. It can be seen that the means differ in the late layer. Note different ranges on $y$ axes. }

    \label{fig:activation_analysis}
\end{figure}

 We compute the means and the standard deviations of the activations in the layers before the first and the last batch normalization layers\footnote{The first batch normalization layer in the Wide Resnet model is inside a residual block, while the last batch normalization layer is before the softmax activation.} for three sets of examples: (i) labeled examples, (ii) unlabeled examples, and (iii) all examples together. Results are shown in Figure~\ref{fig:activation_analysis}. Crucially, the statistics of the labeled examples and the unlabeled examples become significantly different before the last batch normalization. 
 
 
Previous work has found out that batch normalization under-performs when the activation distribution is multimodal~\cite{deecke2019,kalayeh2019}. This and the above analysis suggest that one of the key mechanisms by which Split-BN improves performance is by reducing the negative effect of multimodal distribution of hidden activations on batch normalization layers.


\subsection{Split-BN improves convergence speed}
\label{sec:splitbnconvergencespeed}

In this section we provide a complementary perspective on why Split-BN improves performance. We analyse one of the experiments in Section~\ref{sec:other_distortions} corresponding to training with Mean Teacher and a domain shift introduced by grayscale.
We pick the learning rate that achieves the minimal final auxiliary loss value from $\{0.0003, 0.001, 0.003, 0.01\}$ 

\begin{figure}
\captionsetup[subfigure]{position=top}
    \centering
    \begin{subfigure}[b]{0.49\textwidth}
        \centering
        \includegraphics[width=\textwidth]{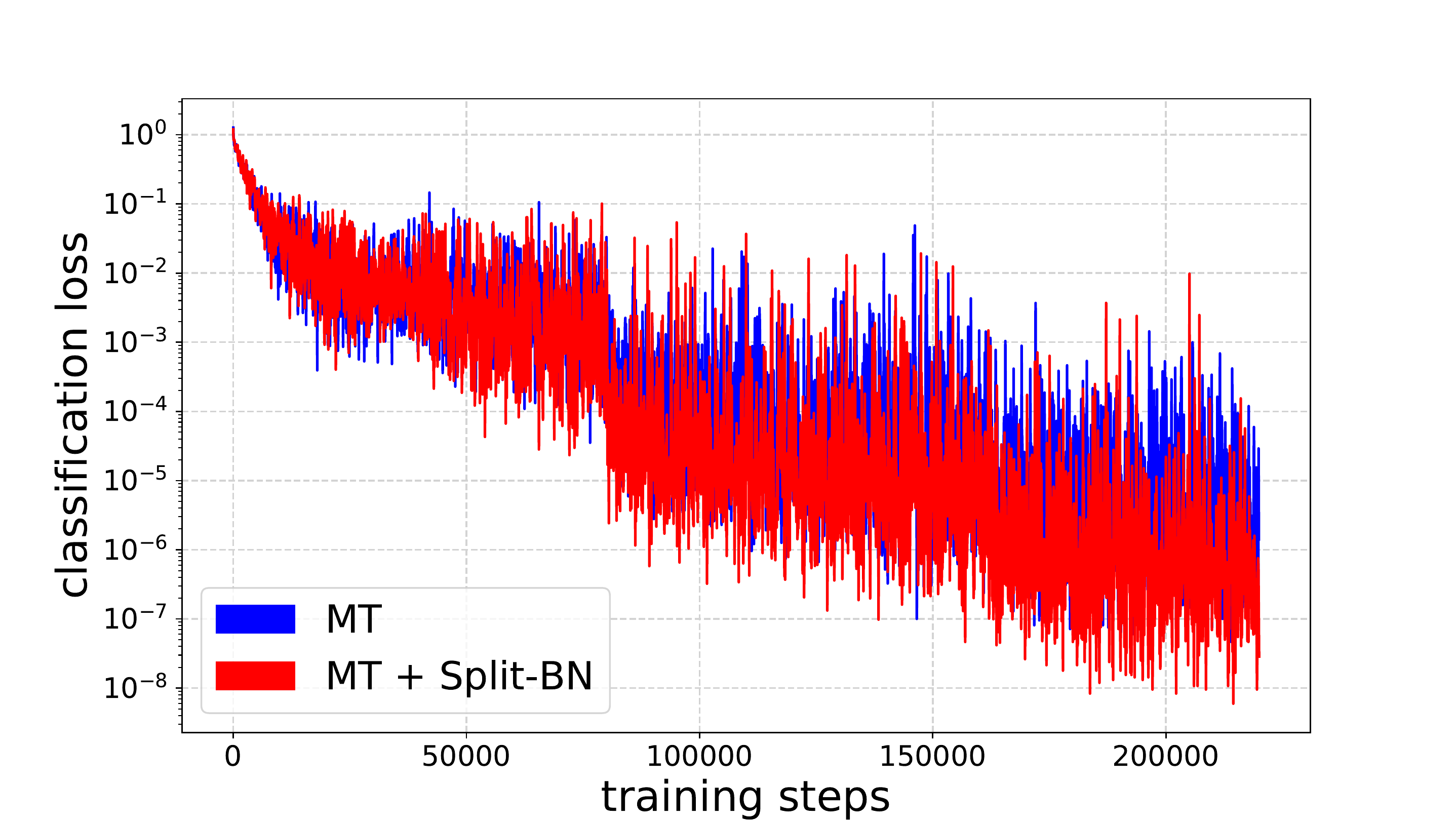}
    \end{subfigure}
    \begin{subfigure}[b]{0.49\textwidth}
        \centering
        \tiny
        \includegraphics[width=\textwidth]{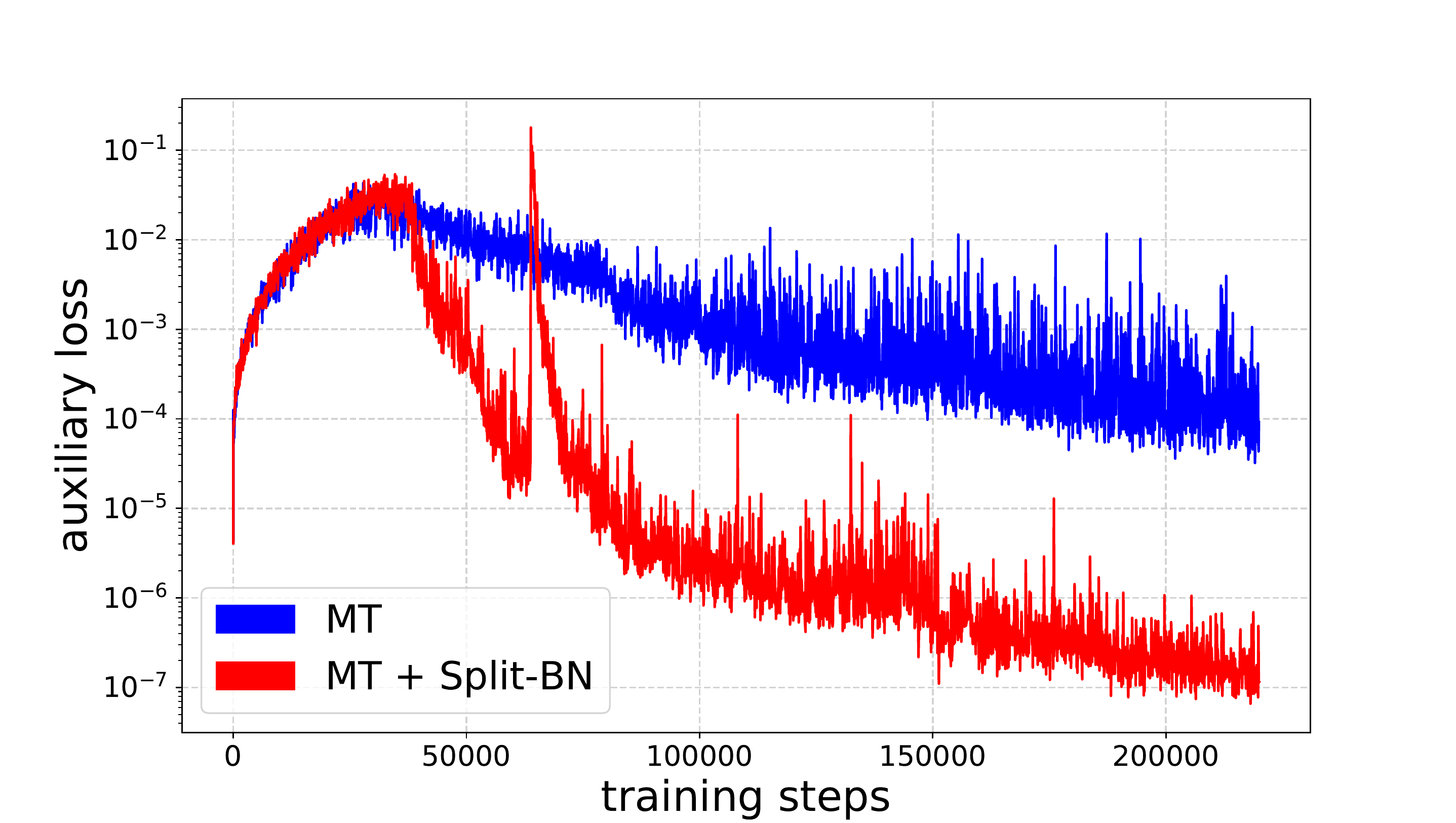}
    \end{subfigure}
    
\caption{Evolution of the classification (left), and the auxiliary losses (right) of Mean Teacher, on a logarithmic scale. Experiment run on the Wide ResNet model from \cite{oliver2018}, and the CIFAR-10 dataset. Split-BN manages to find a lower loss, especially for the auxliary term. Note that the initial increase of the auxiliary loss comes from a warmup schedule. 
}
\label{fig:losses_analysis}
\end{figure}

We first show the evolution of the individual terms (the classification, and the auxiliary) losses in Figure~\ref{fig:losses_analysis}. We can observe that Split-BN enables reaching even an order of magnitude lower final auxiliary loss. This also suggests that further improvements might come from applying more sophisticated optimization techniques.




\subsection{Is Batch Normalization the culprit?}
\label{sec:isbnculprit}

A natural question is whether the lack of robustness of SSL methods simply comes from using batch normalization layers. We repeat the experiments from Section~\ref{sec:other_distortions} using 
a model from~\cite{Miyato2018VirtualAT} and~\cite{tarvainen2017}, where we removed all batch normalization layers. See Appendix~\ref{app:experimental_details} for experimental details.

\begin{table}
    \caption{Repeated experiments from Table~\ref{tab:distortions} using a model from ~\cite{Miyato2018VirtualAT} and ~\cite{tarvainen2017} where we removed all batch normalization layers. Here also we can see a visible performance drop of SSL under domain shift, which shows that batch normalization layers are not a direct cause of the SSL methods instabily under domain shift..}
  \label{tab:distortions_nobn}
\centering
\setlength\tabcolsep{1.5pt}
\begin{tabular}{>{\scriptsize}l|>{\scriptsize}c||>{\scriptsize}c||>{\scriptsize}c}
 \toprule
& \textbf{Supervised} & \textbf{MT}  & \textbf{VAT} \\
\midrule
 None  &  \multirow{8}{*}{73.1 $\pm$ 0.7} & 77.5 $\pm$ 0.2  &  74.5 $\pm$ 0.4 \\
 Grayscale  &&  73.8 $\pm$ 0.4  &  73.5 $\pm$ 0.2 \\
 Uniform noise  &&  73.4 $\pm$ 0.1  &  73.7 $\pm$ 0.1 \\
 Salt-and-pepper  &&  73.2 $\pm$ 0.1  &  74.2 $\pm$ 0.4 \\
 Inverted colors  &&  73.2 $\pm$ 0.2  &  69.9 $\pm$ 0.2 \\
 Rotation $90^{\circ}$  &&  72.8 $\pm$ 0.5  &  70.0 $\pm$ 0.5 \\
 Random contrast  &&  76.2 $\pm$ 0.2  &  74.5 $\pm$ 0.5 \\
 Occlusion &&  73.7 $\pm$ 0.4  &  73.1 $\pm$ 0.5 \\
 \bottomrule
\end{tabular}
\end{table}

\begin{figure}
\captionsetup[subfigure]{position=top}
    \centering
    \begin{subfigure}[b]{0.49\textwidth}
        \centering
        \tiny
        \includegraphics[width=\textwidth]{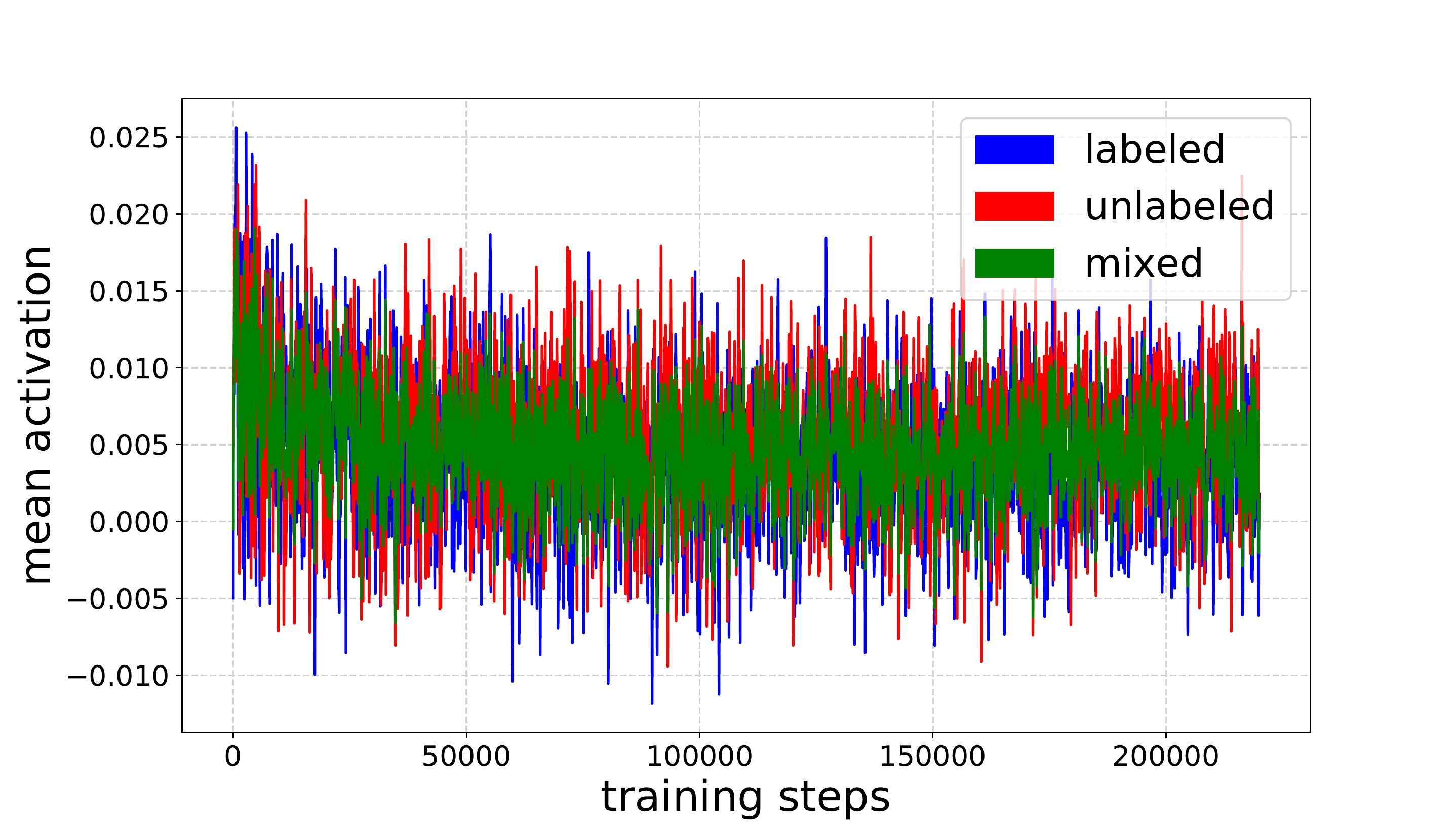}
        \caption{ConvNet with BN, early layer}
    \end{subfigure}
    \begin{subfigure}[b]{0.49\textwidth}
        \centering
        \includegraphics[width=\textwidth]{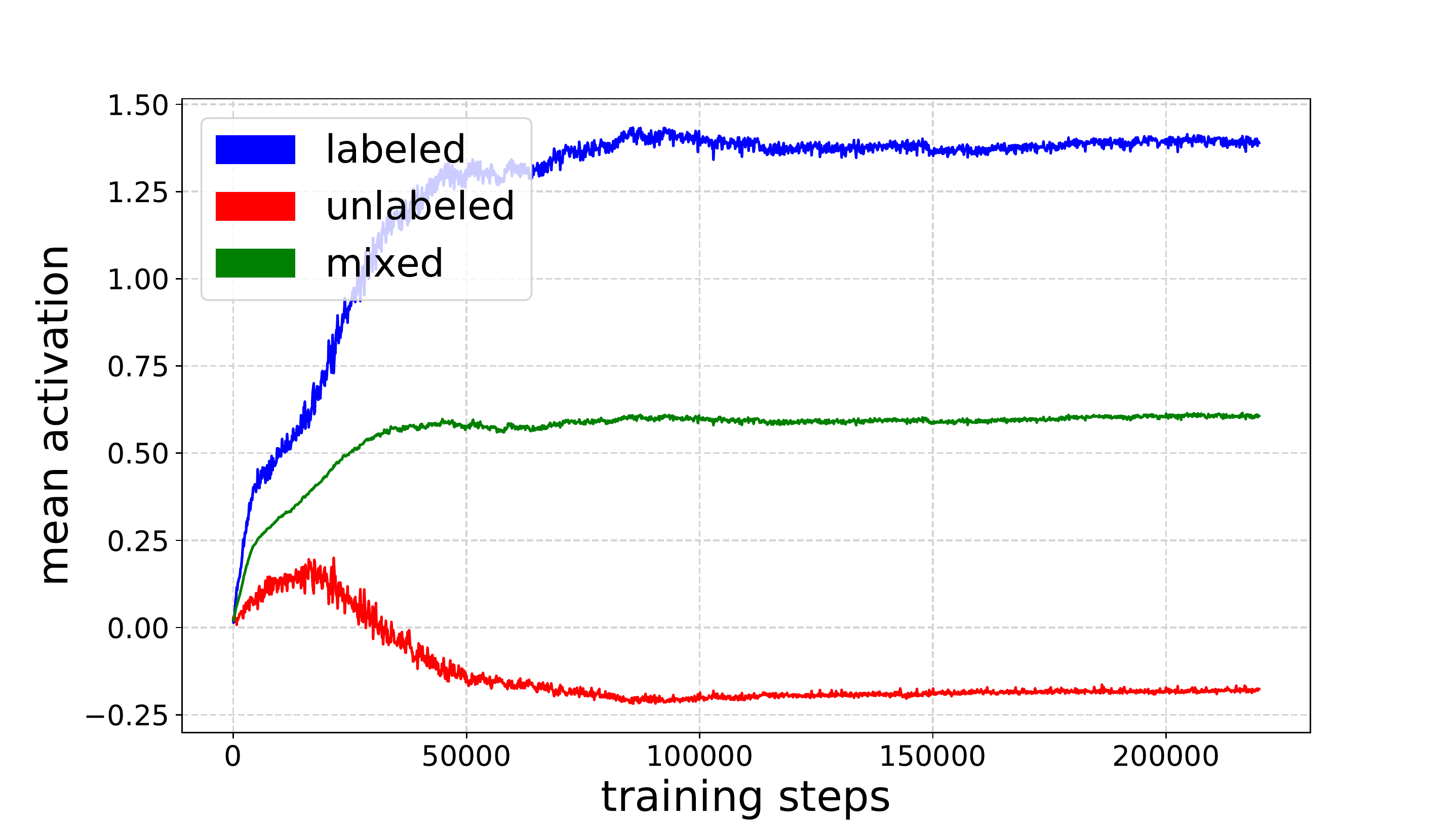}
        \caption{ConvNet with BN, late layer}
    \end{subfigure}

    \begin{subfigure}[b]{0.49\textwidth}
        \centering
        \tiny
        \includegraphics[width=\textwidth]{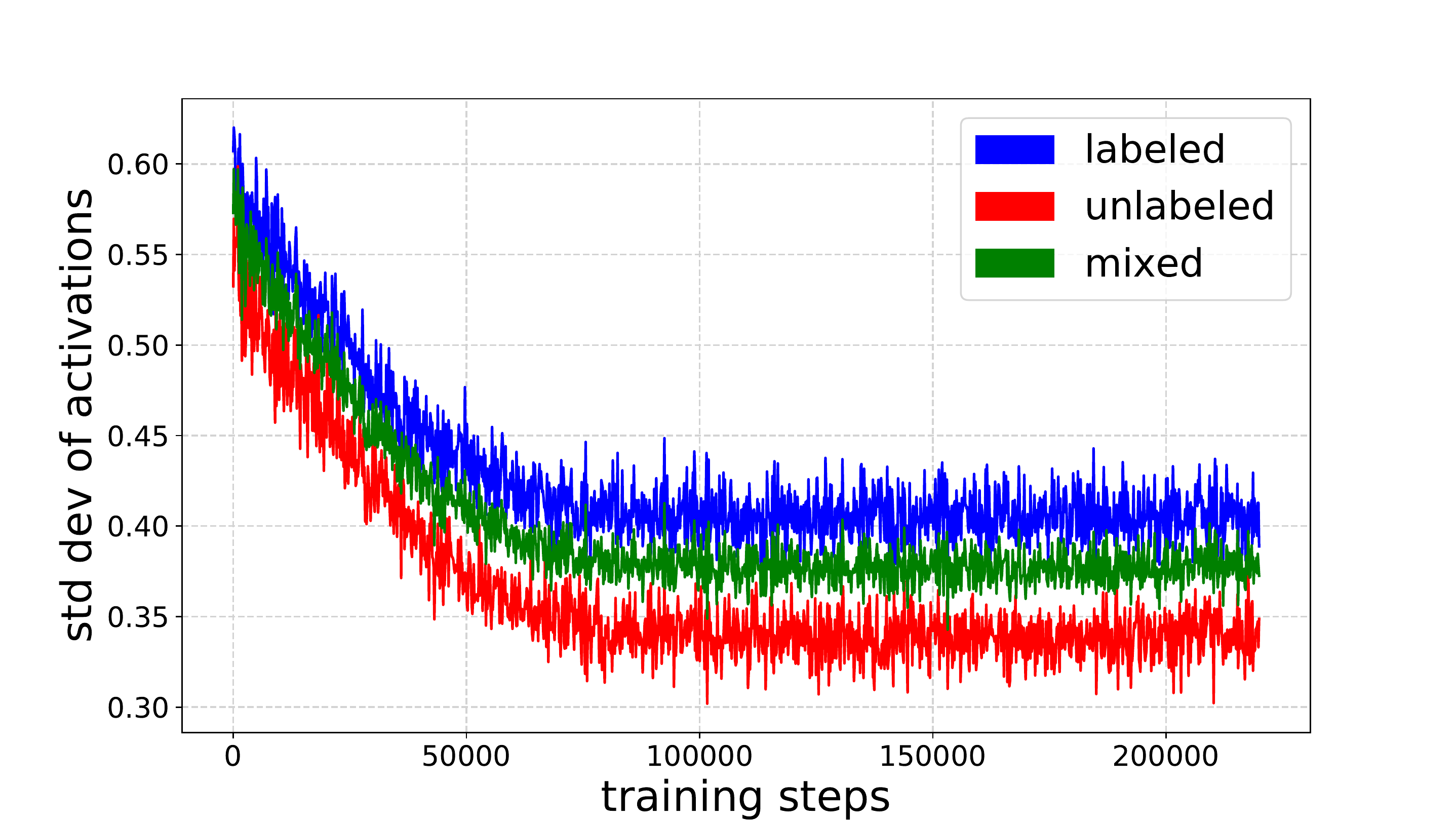}
        \caption{ConvNet with BN, early layer}
    \end{subfigure}
    \begin{subfigure}[b]{0.49\textwidth}
        \centering
        \includegraphics[width=\textwidth]{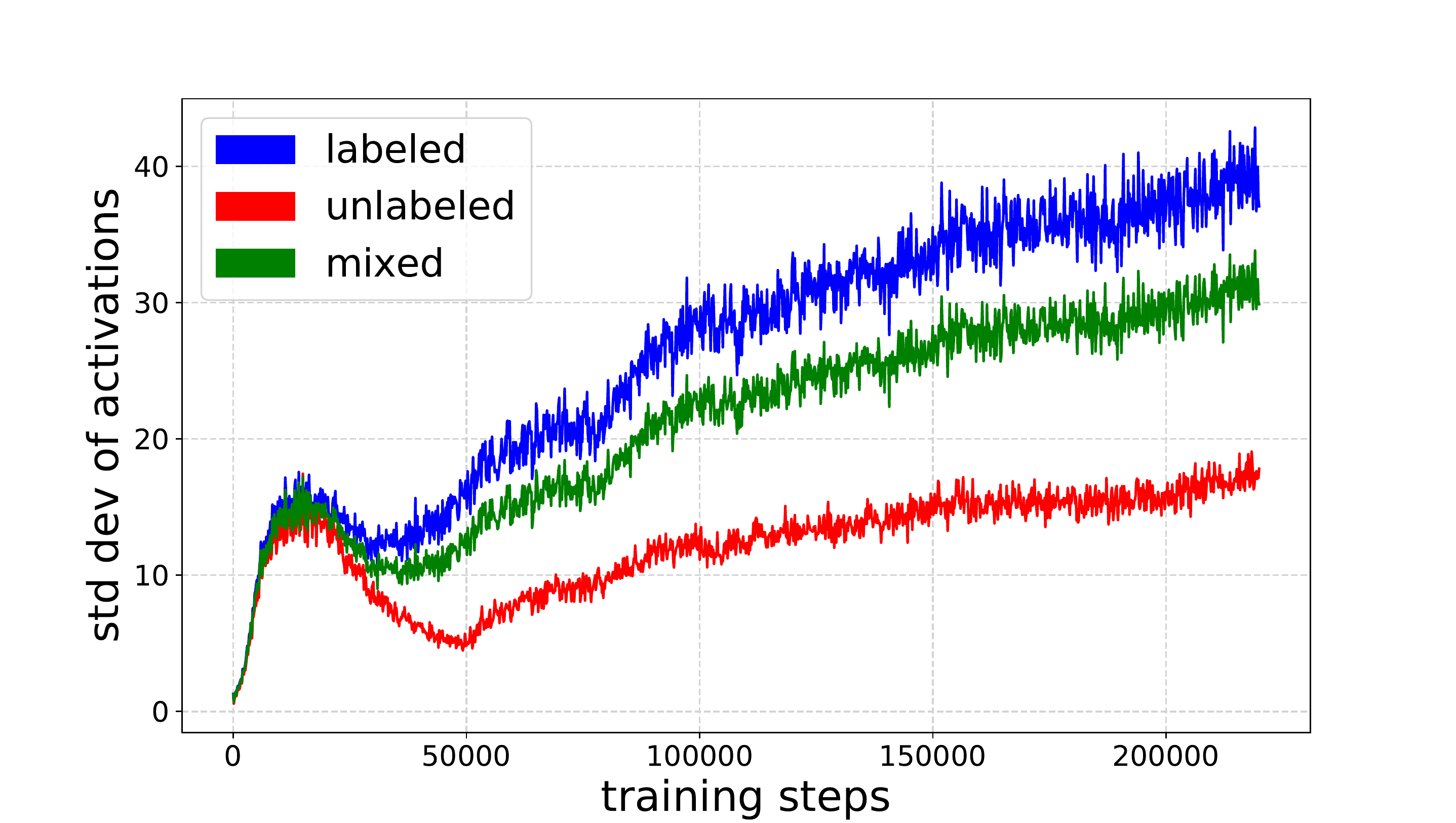}
        \caption{ConvNet with BN, late layer}
    \end{subfigure}

    \begin{subfigure}[b]{0.49\textwidth}
        \centering
        \tiny
        \includegraphics[width=\textwidth]{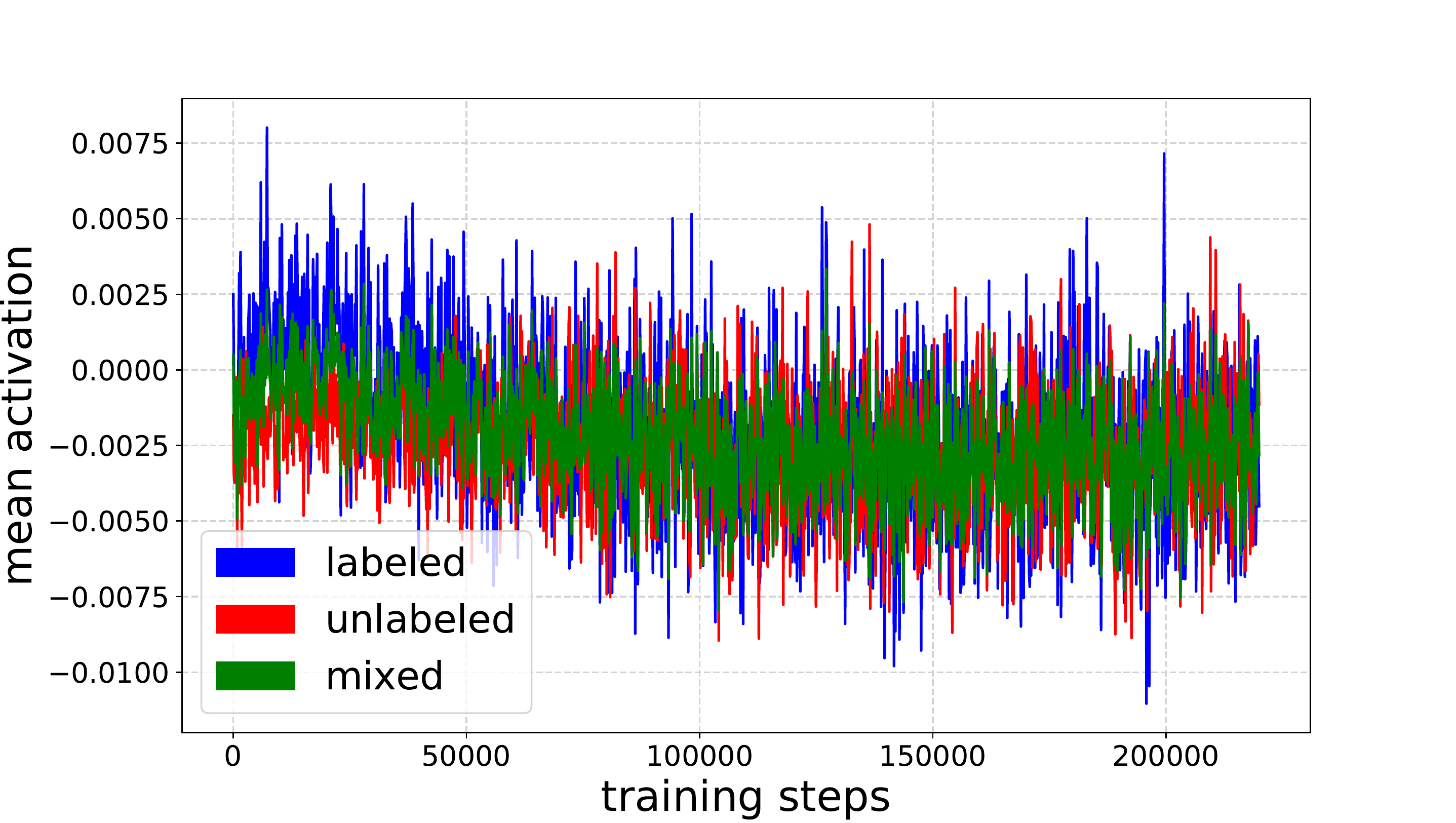}
        \caption{ConvNet without BN, early layer}
    \end{subfigure}
    \begin{subfigure}[b]{0.49\textwidth}
        \centering
        \includegraphics[width=\textwidth]{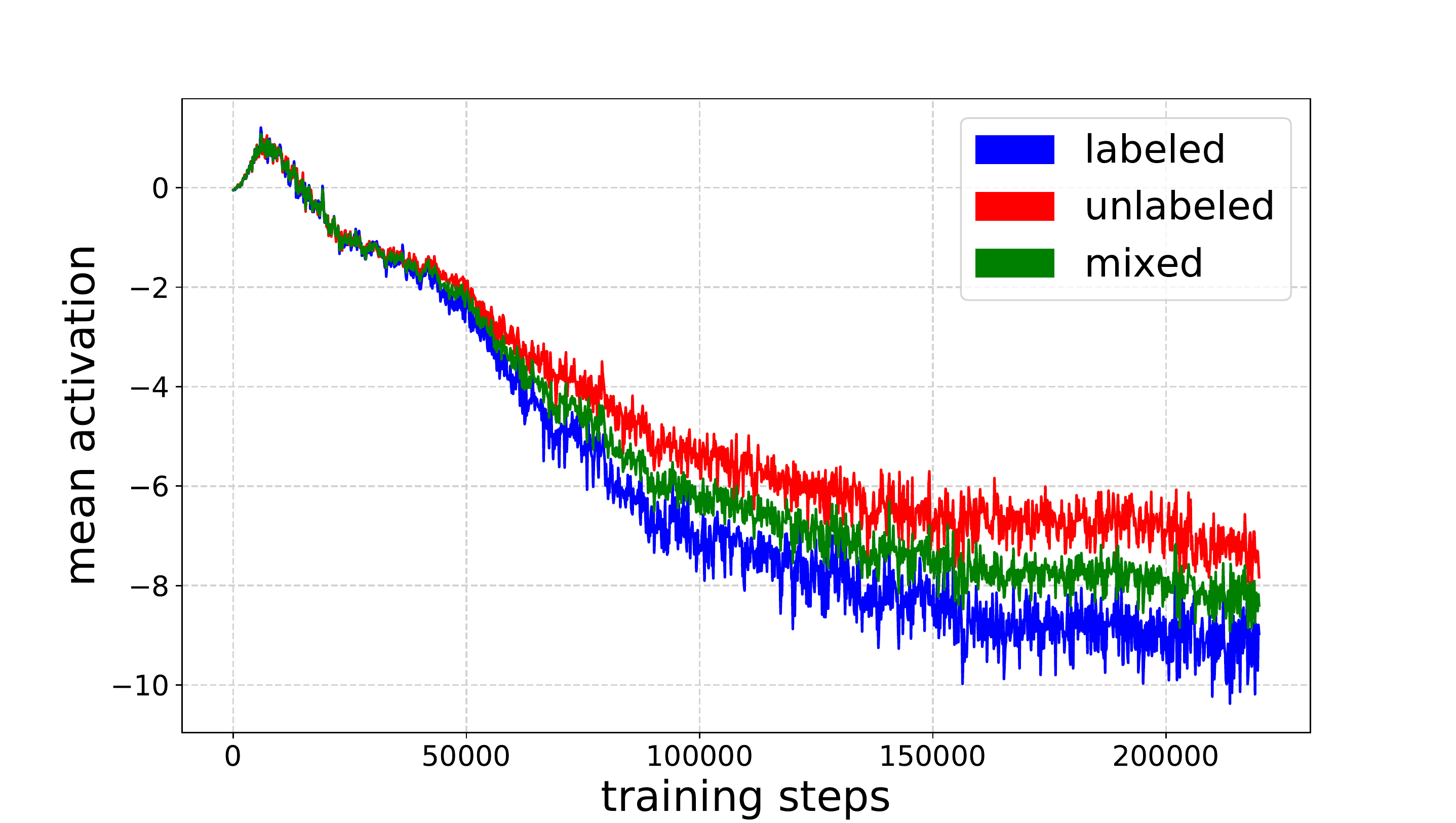}
        \caption{ConvNet without BN, late layer}
    \end{subfigure}

    \begin{subfigure}[b]{0.49\textwidth}
        \centering
        \tiny
        \includegraphics[width=\textwidth]{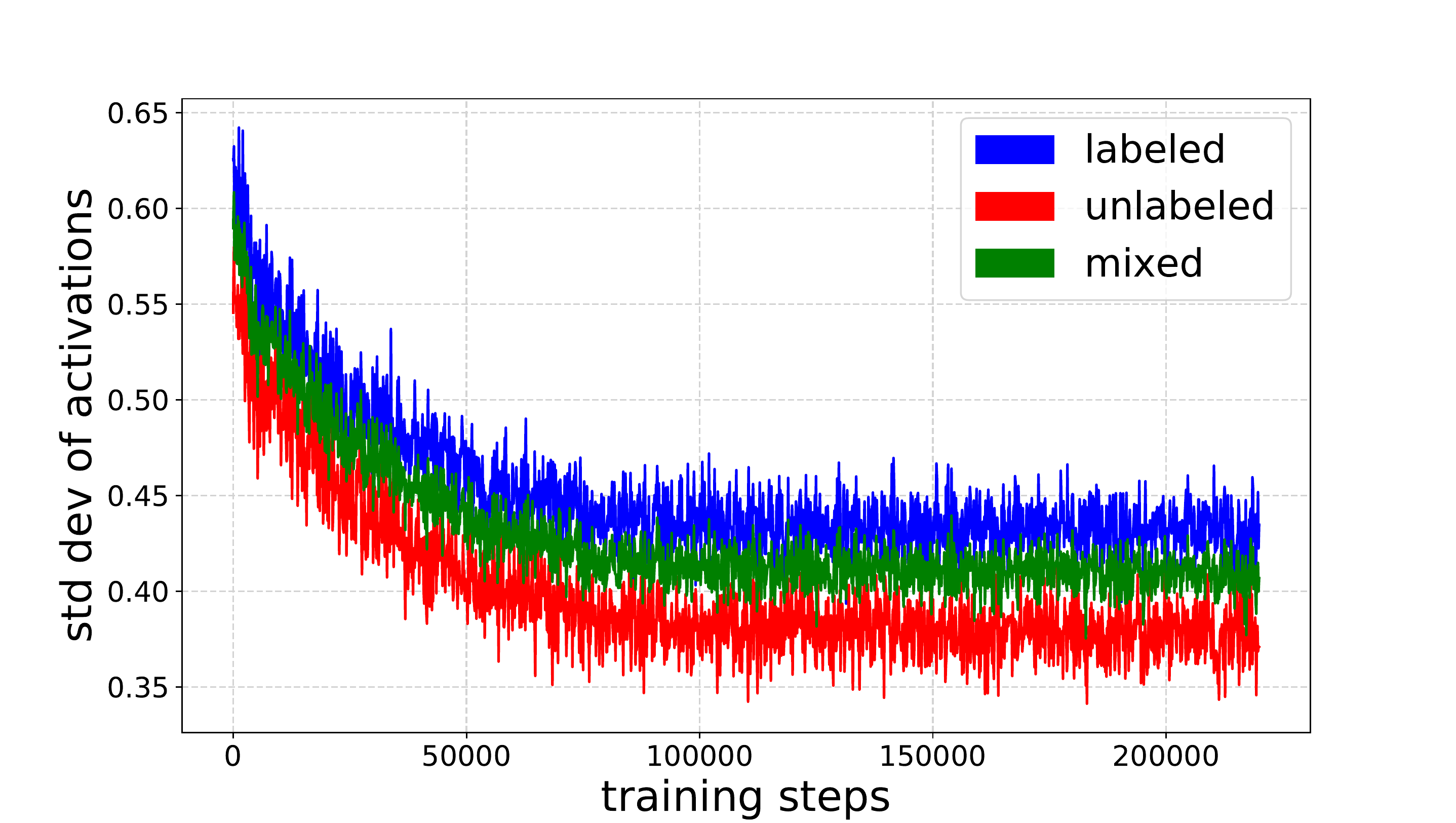}
        \caption{ConvNet without BN, early layer}
    \end{subfigure}
    \begin{subfigure}[b]{0.49\textwidth}
        \centering
        \includegraphics[width=\textwidth]{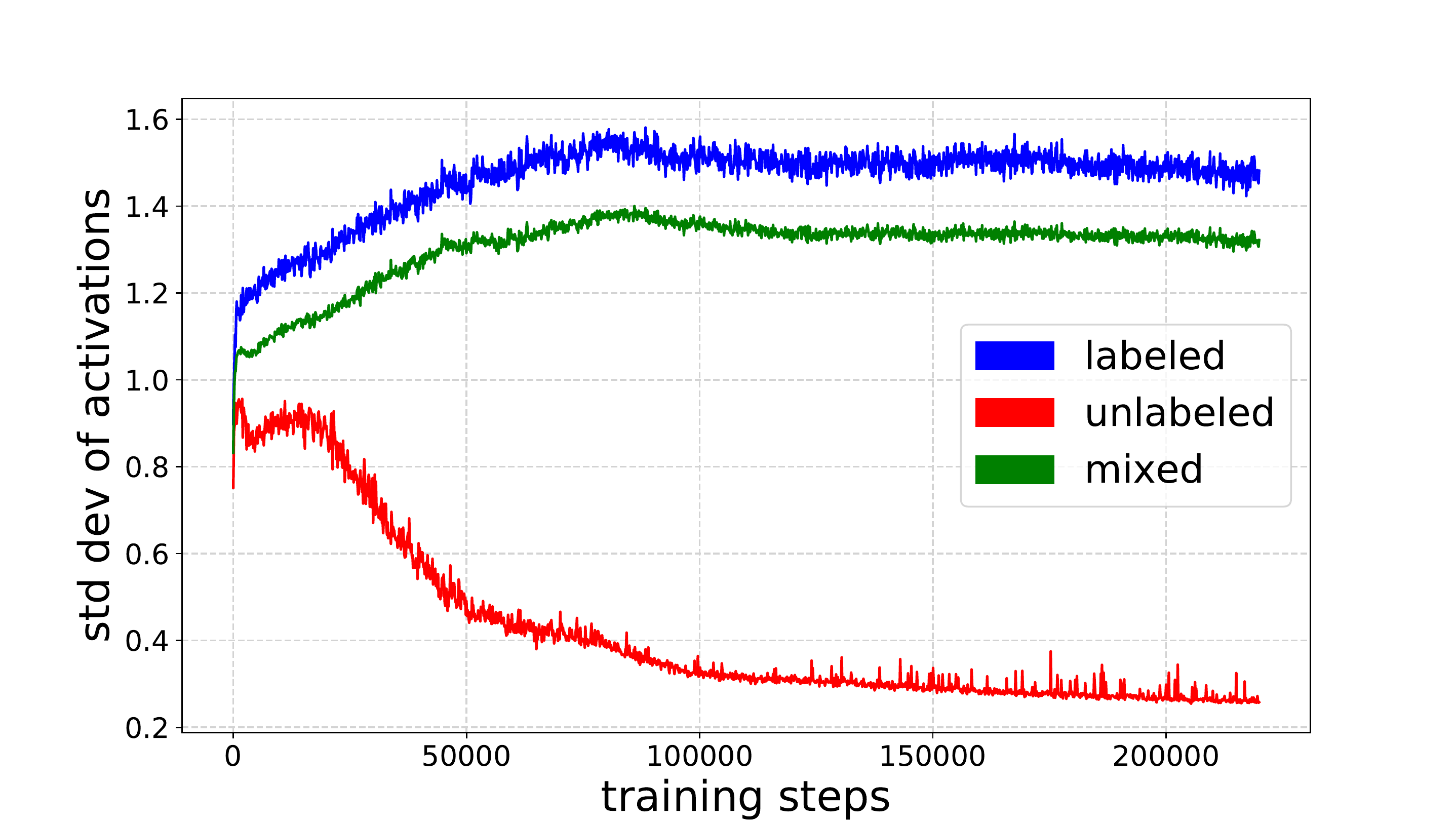}
        \caption{ConvNet without BN, late layer}
    \end{subfigure}
    
    \caption{Means and standard deviations of activations before the first batch normalization layer, and the last batch normalization layer for two different architectures: ConvNet (as in~\cite{Miyato2018VirtualAT} and~\cite{tarvainen2017}) and the same model with removed all batch normalization layers}.
    

\label{fig:activation_analysis_big}
\end{figure}

Table~\ref{tab:distortions_nobn} reports the results. We found that SSL performance may degrade under domain shift also in the case of architectures without batch normalization. This suggests that batch normalization is not the root cause of instability of SSL learning methods.

To further confirm that batch normalization layers are not the root cause, we repeat the analysis from Section~\ref{subsec:activations_analysis} for the model from~\cite{Miyato2018VirtualAT} and~\cite{tarvainen2017} in two variants: with or without the batch normalization layers\footnote{As a technical detail, we replaced Leaky ReLU activation functions with ReLU activation functions.}. In each case we measure the mean and the standard deviation of hidden activations before the first and the last batch normalization layer. As Figure~\ref{fig:activation_analysis_big} shows, regardless whether there is a batch normalization layers, the statistics diverge significantly in the late layer.

\section{Conclusions}

In this work we studied how to use semi-supervised learning if the unlabeled and labeled examples do not come from the same distribution. We provide a simple recommendation: \emph{if a neural network with batch normalization layers is used, the normalization statistics should be computed separately for the unlabeled and labeled data.}
This leads to significant gains in performance; typically canceling out the negative effect of domain shift. 

We also conducted a set of experiments to prove that the effect of domain shift heavily depends on the dataset, and the relationship between distributions for the labeled and the unlabeled data. At one extreme, we were able to engineer a subset of Imagenet dataset (8A8O-ImageNet) where no method was able to leverage the unlabeled examples. However, intuively speaking, there should be a significant amount of useful information in the unlabeled examples, even if they come from a shifted distribution. Future work should look further into improving semi-supervised learning robustness.

\bibliographystyle{splncs04}
\bibliography{main}

\begin{thebibliography}{10}
\providecommand{\url}[1]{\texttt{#1}}
\providecommand{\urlprefix}{URL }
\providecommand{\doi}[1]{https://doi.org/#1}

\bibitem{arpit2017closer}
Arpit, D., Jastrz{\k{e}}bski, S., Ballas, N., Krueger, D., Bengio, E., Kanwal,
  M.S., Maharaj, T., Fischer, A., Courville, A., Bengio, Y., et~al.: A closer
  look at memorization in deep networks. In: Proceedings of the 34th
  International Conference on Machine Learning-Volume 70. pp. 233--242. JMLR.
  org (2017)

\bibitem{ben2010theory}
Ben-David, S., Blitzer, J., Crammer, K., Kulesza, A., Pereira, F., Vaughan,
  J.W.: A theory of learning from different domains. Machine learning
  \textbf{79}(1-2),  151--175 (2010)

\bibitem{Chapelle2010}
Chapelle, O., Schlkopf, B., Zien, A.: Semi-Supervised Learning. The MIT Press,
  1st edn. (2010)

\bibitem{Chen2018}
Chen, K., Yao, L., Zhang, D., Chang, X., Long, G., Wang, S.: Distributionally
  robust semi-supervised learning for people-centric sensing. CoRR
  \textbf{abs/1811.05299} (2018)

\bibitem{chrabaszcz17}
Chrabaszcz, P., Loshchilov, I., Hutter, F.: A downsampled variant of imagenet
  as an alternative to the {CIFAR} datasets. CoRR  \textbf{abs/1707.08819}
  (2017), \url{http://arxiv.org/abs/1707.08819}

\bibitem{Cozman2003}
Cozman, F.G., Cohen, I., Cirelo, M.C.: Semi-supervised learning of mixture
  models. In: Proceedings of the Twentieth International Conference on
  International Conference on Machine Learning. pp. 99--106. ICML'03, AAAI
  Press (2003), \url{http://dl.acm.org/citation.cfm?id=3041838.3041851}

\bibitem{deecke2019}
Deecke, L., Murray, I., Bilen, H.: Mode normalization. In: International
  Conference on Learning Representations (2019),
  \url{https://openreview.net/forum?id=HyN-M2Rctm}

\bibitem{imagenet_cvpr09}
Deng, J., Dong, W., Socher, R., Li, L.J., Li, K., Fei-Fei, L.: {ImageNet: A
  Large-Scale Hierarchical Image Database}. In: CVPR09 (2009)

\bibitem{ganin2016domain}
Ganin, Y., Ustinova, E., Ajakan, H., Germain, P., Larochelle, H., Laviolette,
  F., Marchand, M., Lempitsky, V.: Domain-adversarial training of neural
  networks. The Journal of Machine Learning Research  \textbf{17}(1),
  2096--2030 (2016)

\bibitem{geirhos2018imagenettrained}
Geirhos, R., Rubisch, P., Michaelis, C., Bethge, M., Wichmann, F.A., Brendel,
  W.: Imagenet-trained {CNN}s are biased towards texture; increasing shape bias
  improves accuracy and robustness. In: International Conference on Learning
  Representations (2019), \url{https://openreview.net/forum?id=Bygh9j09KX}

\bibitem{Goodfellow-et-al-2016}
Goodfellow, I., Bengio, Y., Courville, A.: Deep Learning. MIT Press (2016),
  \url{http://www.deeplearningbook.org}

\bibitem{Ioffe2015}
Ioffe, S., Szegedy, C.: Batch normalization: Accelerating deep network training
  by reducing internal covariate shift. In: Proceedings of the 32Nd
  International Conference on International Conference on Machine Learning -
  Volume 37. pp. 448--456. ICML'15, JMLR.org (2015),
  \url{http://dl.acm.org/citation.cfm?id=3045118.3045167}

\bibitem{jo2017}
Jo, J., Bengio, Y.: Measuring the tendency of cnns to learn surface statistical
  regularities. CoRR  \textbf{abs/1711.11561} (2017)

\bibitem{kalayeh2019}
Kalayeh, M., Shah, M.: Training faster by separating modes of variation in
  batch-normalized models. IEEE transactions on pattern analysis and machine
  intelligence  (2019)

\bibitem{laine2016temporal}
Laine, S., Aila, T.: Temporal ensembling for semi-supervised learning. arXiv
  preprint arXiv:1610.02242  (2016)

\bibitem{Li2016}
Li, Y., Wang, N., Shi, J., Liu, J., Hou, X.: Revisiting batch normalization for
  practical domain adaptation. CoRR  \textbf{abs/1603.04779} (2016),
  \url{http://arxiv.org/abs/1603.04779}

\bibitem{liu2018robust}
Liu, X., Zachariah, D., W{\aa}gberg, J.: Robust semi-supervised learning when
  labels are missing at random. arXiv preprint arXiv:1811.10947  (2018)

\bibitem{Miyato2018VirtualAT}
Miyato, T., Maeda, S.i., Ishii, S., Koyama, M.: Virtual adversarial training: a
  regularization method for supervised and semi-supervised learning. IEEE
  transactions on pattern analysis and machine intelligence  (2018)

\bibitem{oliver2018}
Oliver, A., Odena, A., Raffel, C.A., Cubuk, E.D., Goodfellow, I.: Realistic
  evaluation of deep semi-supervised learning algorithms. In: Advances in
  Neural Information Processing Systems. pp. 3239--3250 (2018)

\bibitem{nasim2018}
{Rahaman}, N., {Baratin}, A., {Arpit}, D., {Draxler}, F., {Lin}, M.,
  {Hamprecht}, F.A., {Bengio}, Y., {Courville}, A.: {On the Spectral Bias of
  Neural Networks}. arXiv e-prints arXiv:1806.08734 (Jun 2018)

\bibitem{ruder2018strong}
Ruder, S., Plank, B.: Strong baselines for neural semi-supervised learning
  under domain shift. arXiv preprint arXiv:1804.09530  (2018)

\bibitem{tarvainen2017}
Tarvainen, A., Valpola, H.: Mean teachers are better role models:
  Weight-averaged consistency targets improve semi-supervised deep learning
  results. In: Advances in neural information processing systems. pp.
  1195--1204 (2017)

\bibitem{wang2018learning}
Wang, H., He, Z., Xing, E.P.: Learning robust representations by projecting
  superficial statistics out. In: International Conference on Learning
  Representations (2019), \url{https://openreview.net/forum?id=rJEjjoR9K7}

\bibitem{Zagoruyko2016WRN}
Zagoruyko, S., Komodakis, N.: Wide residual networks. In: BMVC (2016)

\end{thebibliography}


\newpage

\appendix
%

\section{Details of experimental setting}
\label{app:experimental_details}

We largely use implementation of the architecture and SSL methods from \cite{oliver2018}.

In all experiments we use the following: batch size equal to 100; Adam optimizer with all hyperparameters but learning rate set to default. In supervised data, there is 400 images per class. Every number we report is accuracy on test set computed using checkpoint with the highest validation accuracy during the training. Every experiment is repeated 2 times.

In experiments from Tables \ref{tab:cifar10_class_mismatch}, \ref{tab:imagenet_class_mismatch},
\ref{tab:our_imagenet_class_mismatch}, \ref{tab:distortions},
\ref{tab:distortions_our_imagenet}, we use Wide ResNet architecture (WRN-28-2 variant with depth 28 and width 2) and all Mean Teacher and VAT parameters as in \cite{oliver2018}.

\paragraph{Experiments from Table \ref{tab:cifar10_class_mismatch}} We use the experimental setting from \cite{oliver2018}. The classification is performed on 6 animal classes from CIFAR-10. Additional unsupervised data contains 4 classes and has a varied degree of class mismatch with supervised data -- from $0\%$ (no mismatch) to $100\%$ (completely different classes). CIFAR-10 data is preprocessed using global contrast normalization and ZCA normalization. Data is augmented with random horizontal flips, random translation by up to 2 pixels, and Gaussian
input noise with standard deviation 0.15. Number of steps is 500000. Learning rate is as in \cite{oliver2018}: for Mean Teacher, initial learning rate is 0.0004, decaying by 0.2 after 400000 steps; for VAT, initial learning rate is 0.003, decaying by 0.2 after 400000 steps. Auxiliary SSL loss is warmed-up for 200000 steps.

\paragraph{Experiments from Table \ref{tab:imagenet_class_mismatch}} The classification is performed on 20 random classes from ImageNet rescaled do 32x32. Additional unsupervised data contains 20 classes and has a varied degree of class mismatch with supervised data -- from $0\%$ (no mismatch) to $100\%$ (completely different classes). Data is augmented with random horizontal flips. Learning rate is as in \cite{oliver2018}: initial learning rate is 0.0004, decaying by 0.2 after 400000 steps. Auxiliary SSL loss is warmed-up for 200000 steps.

\paragraph{Experiments from Table \ref{tab:our_imagenet_class_mismatch}} We use 8 animal classes from 8A8O-ImageNet as supervised data. Additional unsupervised data contains 8 classes and has a varied degree of class mismatch with supervised data -- from $0\%$ (no mismatch) to $100\%$ (completely different classes). Data is augmented with random horizontal flips, random translation by up to 2 pixels, and gaussian noise. For each experiment, we search for optimal learning rate in the set $\{0.0003, 0.001, 0.003, 0.01\}$. The training lasts for 170000 steps. Learning rate decays by 0.2 after 100000 steps. Auxiliary SSL loss coefficient is warmed-up for 50000 steps to reach $8.0$ for Mean Teacher and $0.3$ for VAT.

\paragraph{Experiments from Table \ref{tab:distortions}} Here we use CIFAR-10 without preprocessing. Data is augmented with random horizontal flips and random translation by up to 2 pixels. For each experiment, we search for optimal learning rate in the set $\{0.0003, 0.001, 0.003, 0.01\}$. The training lasts for 170000 steps. Learning rate decays by 0.2 after 100000 steps. Auxiliary SSL loss coefficient is warmed-up for 50000 steps to reach $8.0$ for Mean Teacher and $0.3$ for VAT.

\paragraph{Experiments from Table \ref{tab:distortions_our_imagenet}} We use 8 animal classes from 8A8O-ImageNet. Data is augmented with random horizontal flips, random translation by up to 2 pixels, and gaussian noise. For each experiment, we search for optimal learning rate in the set $\{0.0003, 0.001, 0.003, 0.01\}$. The training lasts for 170000 steps. Learning rate decays by 0.2 after 100000 steps. Auxiliary SSL loss coefficient is warmed-up for 50000 steps to reach $8.0$ for Mean Teacher and $0.3$ for VAT.

\paragraph{Experiments from Table \ref{tab:distortions_nobn}} We use ConvNet architecture from~\cite{Miyato2018VirtualAT} and~\cite{tarvainen2017} with batch normalization eliminated -- see Table~\ref{tab:convnet} for details. We use CIFAR-10 without preprocessing. Data is augmented with gaussian noise with standard deviation 0.15, random horizontal flips and random translation by up to 2 pixels. For each experiment, we search for optimal learning rate in the set $\{0.0001, 0.0003, 0.001\}$. The training lasts for 220000 steps. Learning rate decays by 0.25 after every 80000 steps. Auxiliary SSL loss coefficient is warmed-up for 50000 steps to reach $8.0$ for Mean Teacher and $0.1$ for VAT.

\begin{table}
\centering
\caption{ConvNet architecture, as used in~\cite{Miyato2018VirtualAT} and~\cite{tarvainen2017}, but without batch normalization.}
\begin{tabular}{ >{\scriptsize}c }
\toprule
convolutional, $128$ filters, $3\times3$, same padding, leaky ReLU \\
convolutional, $128$ filters, $3\times3$, same padding, leaky ReLU \\
convolutional, $128$ filters, $3\times3$, same padding, leaky ReLU \\
max pooling $2\times2$ \\
dropout   $p=0.5$ \\
\midrule
convolutional,  $256$ filters, $3\times3$, same padding, leaky ReLU \\
convolutional,  $256$ filters, $3\times3$, same padding, leaky ReLU \\
convolutional,  $256$ filters, $3\times3$, same padding, leaky ReLU \\
max pooling $2\times2$ \\
dropout  $p=0.5$ \\
\midrule
convolutional,  $512$ filters, $3\times3$, valid padding, leaky ReLU \\
convolutional,  $256$ filters, $1\times1$, same padding, leaky ReLU \\
convolutional,  $128$ filters, $1\times1$, same padding, leaky ReLU \\
\midrule
average pooling ($6\times6 \to 1\times$1) \\
fully connected, $128 \to 10$ \\
softmax \\
\bottomrule
\end{tabular}
\label{tab:convnet}
\end{table}

\section{Details of 8A8O-ImageNet dataset}
\label{app:8a8o}
We created each of the 8A8O-ImageNet classes by gathering several ImageNet classes -- for details see Table~\ref{tab:8a8o_classes}. After merging those ImageNet classes, we randomly chose subset of examples so that for every 8A8O-ImageNet class, there are 4800 examples in the training set, 400 in the validation set and 200 in the test set. The first 8 classes are various animals and the other are non-animal.

We used 32x32 images as in \cite{chrabaszcz17}.

\begin{table}
\centering
\caption{Original ImageNet classes included in 8A8O-ImageNet classes.}
\begin{tabular}{>{\scriptsize}l|>{\scriptsize}p{0.8\textwidth} }
\toprule
class & included ImageNet classes \\
\midrule
\texttt{bear} & \texttt{n02132136}, \texttt{n02133161}, \texttt{n02134084}, \texttt{n02134418}\\
\midrule
\texttt{cat} & \texttt{n02122878}, \texttt{n02123045}, \texttt{n02123159}, \texttt{n02126465}, \texttt{n02123394}, \texttt{n02123597}, \texttt{n02124075}, \texttt{n02125311}\\
\midrule
\texttt{bird} & \texttt{n01321123}, \texttt{n01514859}, \texttt{n01792640}, \texttt{n07646067}, \texttt{n01530575}, \texttt{n01531178}, \texttt{n01532829}, \texttt{n01534433}, \texttt{n01537544}, \texttt{n01558993}, \texttt{n01562265}, \texttt{n01560419}, \texttt{n01582220}, \texttt{n10281276}, \texttt{n01592084}, \texttt{n01601694}, \texttt{n01614925}, \texttt{n01616318}, \texttt{n01622779}, \texttt{n01795545}, \texttt{n01796340}, \texttt{n01797886}, \texttt{n01798484}, \texttt{n01817953}, \texttt{n01818515}, \texttt{n01819313}, \texttt{n01820546}, \texttt{n01824575}, \texttt{n01828970}, \texttt{n01829413}, \texttt{n01833805}, \texttt{n01843065}, \texttt{n01843383}, \texttt{n01855032}, \texttt{n01855672}, \texttt{n07646821}, \texttt{n01860187}, \texttt{n02002556}, \texttt{n02002724}, \texttt{n02006656}, \texttt{n02007558}, \texttt{n02009229}, \texttt{n02009912}, \texttt{n02011460}, \texttt{n02013706}, \texttt{n02017213}, \texttt{n02018207}, \texttt{n02018795}, \texttt{n02025239}, \texttt{n02027492}, \texttt{n02028035}, \texttt{n02033041}, \texttt{n02037110}, \texttt{n02051845}, \texttt{n02056570}\\
\midrule
\texttt{dog} & \texttt{n02085782}, \texttt{n02085936}, \texttt{n02086079}, \texttt{n02086240}, \texttt{n02086646}, \texttt{n02086910}, \texttt{n02087046}, \texttt{n02087394}, \texttt{n02088094}, \texttt{n02088238}, \texttt{n02088364}, \texttt{n02088466}, \texttt{n02088632}, \texttt{n02089078}, \texttt{n02089867}, \texttt{n02089973}, \texttt{n02090379}, \texttt{n02090622}, \texttt{n02090721}, \texttt{n02091032}, \texttt{n02091134}, \texttt{n02091244}, \texttt{n02091467}, \texttt{n02091635}, \texttt{n02091831}, \texttt{n02092002}, \texttt{n02092339}, \texttt{n02093256}, \texttt{n02093428}, \texttt{n02093647}, \texttt{n02093754}, \texttt{n02093859}, \texttt{n02093991}, \texttt{n02094114}, \texttt{n02094258}, \texttt{n02094433}, \texttt{n02095314}, \texttt{n02095570}, \texttt{n02095889}, \texttt{n02096051}, \texttt{n02096294}, \texttt{n02096437}, \texttt{n02096585}, \texttt{n02097047}, \texttt{n02097130}, \texttt{n02097209}, \texttt{n02097298}, \texttt{n02097474}, \texttt{n02097658}, \texttt{n02098105}, \texttt{n02098286}, \texttt{n02099267}, \texttt{n02099429}, \texttt{n02099601}, \texttt{n02099712}, \texttt{n02099849}, \texttt{n02100236}, \texttt{n02100583}, \texttt{n02100735}, \texttt{n02100877}, \texttt{n02101006}, \texttt{n02101388}, \texttt{n02101556}, \texttt{n02102040}, \texttt{n02102177}, \texttt{n02102318}, \texttt{n02102480}, \texttt{n02102973}, \texttt{n02104029}, \texttt{n02104365}, \texttt{n02105056}, \texttt{n02105162}, \texttt{n02105251}, \texttt{n02105505}, \texttt{n02105641}, \texttt{n02105855}, \texttt{n02106030}, \texttt{n02106166}, \texttt{n02106382}, \texttt{n02106550}, \texttt{n02106662}, \texttt{n02107142}, \texttt{n02107312}, \texttt{n02107574}, \texttt{n02107683}, \texttt{n02107908}, \texttt{n02108000}, \texttt{n02108422}, \texttt{n02108551}, \texttt{n02108915}, \texttt{n02109047}, \texttt{n02109525}, \texttt{n02109961}, \texttt{n02110063}, \texttt{n02110185}, \texttt{n02110627}, \texttt{n02110806}, \texttt{n02110958}, \texttt{n02111129}, \texttt{n02111277}, \texttt{n08825211}, \texttt{n02111500}, \texttt{n02112018}, \texttt{n02112350}, \texttt{n02112706}, \texttt{n02113023}, \texttt{n02113624}, \texttt{n02113712}, \texttt{n02113799}, \texttt{n02113978}\\
\midrule
\texttt{monkey} & \texttt{n02494079}, \texttt{n02489166}, \texttt{n02493793}, \texttt{n02492660}, \texttt{n02480855}, \texttt{n02481823}, \texttt{n02487347}\\
\midrule
\texttt{spider} & \texttt{n01773157}, \texttt{n01773549}, \texttt{n01773797}, \texttt{n01774384}, \texttt{n01774750}, \texttt{n01775062}\\
\midrule
\texttt{fish} & \texttt{n01443537}, \texttt{n02607072}, \texttt{n02643566}, \texttt{n02526121}, \texttt{n02606052}, \texttt{n02655020}, \texttt{n02640242}\\
\midrule
\texttt{snake} & \texttt{n01728572}, \texttt{n01728920}, \texttt{n01729322}, \texttt{n01729977}, \texttt{n01734418}, \texttt{n01735189}, \texttt{n01737021}\\
\midrule
\texttt{boat} & \texttt{n02951358}, \texttt{n03344393}, \texttt{n03662601}, \texttt{n04273569}, \texttt{n04612373}, \texttt{n04612504}\\
\midrule
\texttt{bottle} & \texttt{n02823428}, \texttt{n03937543}, \texttt{n03983396}, \texttt{n04557648}, \texttt{n04560804}, \texttt{n04579145}, \texttt{n04591713}\\
\midrule
\texttt{truck} & \texttt{n03345487}, \texttt{n03417042}, \texttt{n03770679}, \texttt{n03796401}, \texttt{n00319176}, \texttt{n01016201}, \texttt{n03930630}, \texttt{n03930777}, \texttt{n05061003}, \texttt{n06547832}, \texttt{n10432053}, \texttt{n03977966}, \texttt{n04461696}, \texttt{n04467665}\\
\midrule
\texttt{car} & \texttt{n02814533}, \texttt{n03100240}, \texttt{n03100346}, \texttt{n13419325}, \texttt{n04285008}, \texttt{n03777568}\\
\midrule
\texttt{shoe} & \texttt{n04120489}, \texttt{n03124043}, \texttt{n04133789}, \texttt{n03047690}\\
\midrule
\texttt{string\_instrument} & \texttt{n02992211}, \texttt{n04536866}, \texttt{n02676566}, \texttt{n03272010}\\
\midrule
\texttt{fungus} & \texttt{n12985857}, \texttt{n07734744}, \texttt{n13052670}, \texttt{n13044778}\\
\midrule
\texttt{cloth} & \texttt{n02916936}, \texttt{n04370456}, \texttt{n02963159}, \texttt{n04136333}, \texttt{n03866082}\\
\bottomrule
\end{tabular}
\label{tab:8a8o_classes}
\end{table}

\end{document}